%% file: neurips_2025.tex
\documentclass{article}

\PassOptionsToPackage{square,comma,numbers,sort&compress}{natbib}

\usepackage[final]{neurips_2025}

\usepackage[utf8]{inputenc} 
\usepackage[T1]{fontenc}    
\usepackage{hyperref}       
\usepackage{url}            
\usepackage{booktabs}       
\usepackage{amsfonts}       
\usepackage{amsmath}
\usepackage{nicefrac}       
\usepackage{microtype}      
\usepackage{xcolor}         
\usepackage{xspace}
\usepackage{xfrac}
\usepackage{subcaption}
\usepackage{graphicx}
\usepackage[most]{tcolorbox}
\usepackage{cleveref}
\usepackage{tabularx}
\usepackage{bm} 
\usepackage{multirow}
\usepackage{makecell}
\usepackage{float}

\definecolor{darkgreen}{rgb}{0.09, 0.45, 0.27}
\newcommand{\bestscore}[1]{\textcolor{darkgreen}{\mathbf{#1}}}
\newcommand{\greentext}[1]{\textcolor{darkgreen}{\textbf{#1}}}

\definecolor{carbon-gray-10}{cmyk}{0.0, 0.0, 0.0, 0.04, 1.00}
\definecolor{carbon-gray-90}{cmyk}{0.0, 0.0, 0.0, 0.85, 1.00}

\newcolumntype{L}[1]{>{\raggedright\let\newline\\\arraybackslash\hspace{0pt}}m{#1}}
\newcolumntype{C}[1]{>{\centering\let\newline\\\arraybackslash\hspace{0pt}}m{#1}}
\newcolumntype{R}[1]{>{\raggedleft\let\newline\\\arraybackslash\hspace{0pt}}m{#1}}
\newcolumntype{Y}{>{\centering\arraybackslash}X}

\title{MindForge: Empowering Embodied Agents with Theory of Mind for Lifelong
Cultural Learning}

\newcommand{\collabvoyager}[0]{\mbox{\textsc{MindForge}}\xspace}

\author{%
  Mircea Lic\u{a}\textsuperscript{*}\\
  Delft University of Technology\\
  \texttt{mlica@tudelft.nl} \\
  \And
  Ojas Shirekar\textsuperscript{*} \\
  Delft University of Technology \\
  \texttt{o.k.shirekar@tudelft.nl} \\
  \And
  Baptiste Colle \\
  Delft University of Technology \\
  \texttt{collebaptiste@gmail.com} \\
  \And
  Chirag Raman \\
  Delft University of Technology \\
  \texttt{c.a.raman@tudelft.nl}
}

\begin{document}

\maketitle
\begingroup
\renewcommand\thefootnote{*}
\footnotetext{Equal Contribution.}
\endgroup

\begin{abstract}
Embodied agents powered by large language models (LLMs), such as Voyager, promise open-ended competence in worlds such as Minecraft. However, when powered by open-weight LLMs they still falter on elementary tasks after domain-specific fine-tuning. We propose \collabvoyager, a generative-agent framework for cultural lifelong learning through explicit perspective taking. We introduce three key innovations: (1) a structured theory of mind representation linking percepts, beliefs, desires, and actions; (2) natural inter-agent communication; and (3) a multi-component memory system. Following the cultural learning framework, we test \collabvoyager in both instructive and collaborative settings within Minecraft. In an \textit{instructive} setting with GPT-4, \collabvoyager agents powered by open-weight LLMs significantly outperform their Voyager counterparts in basic tasks yielding $3\times$ more tech-tree milestones and collecting $2.3\times$ more unique items than the Voyager baseline. Furthermore, in fully \textit{collaborative} settings, we find that the performance of two underachieving agents improves with more communication rounds, echoing the Condorcet Jury Theorem. \collabvoyager agents demonstrate sophisticated behaviors, including expert-novice knowledge transfer, collaborative problem solving, and adaptation to out-of-distribution tasks through accumulated cultural experiences. The code can be found at \href{https://github.com/tapri-lab/mindforge}{\textcolor{blue}{https://github.com/tapri-lab/mindforge}}.
\end{abstract}

\begin{figure*}[h]
    \captionsetup[subfigure]{aboveskip=-1pt,belowskip=-1pt}
     \centering
     \begin{subfigure}[b]{0.48\textwidth}
         \centering
         \includegraphics[width=\linewidth]{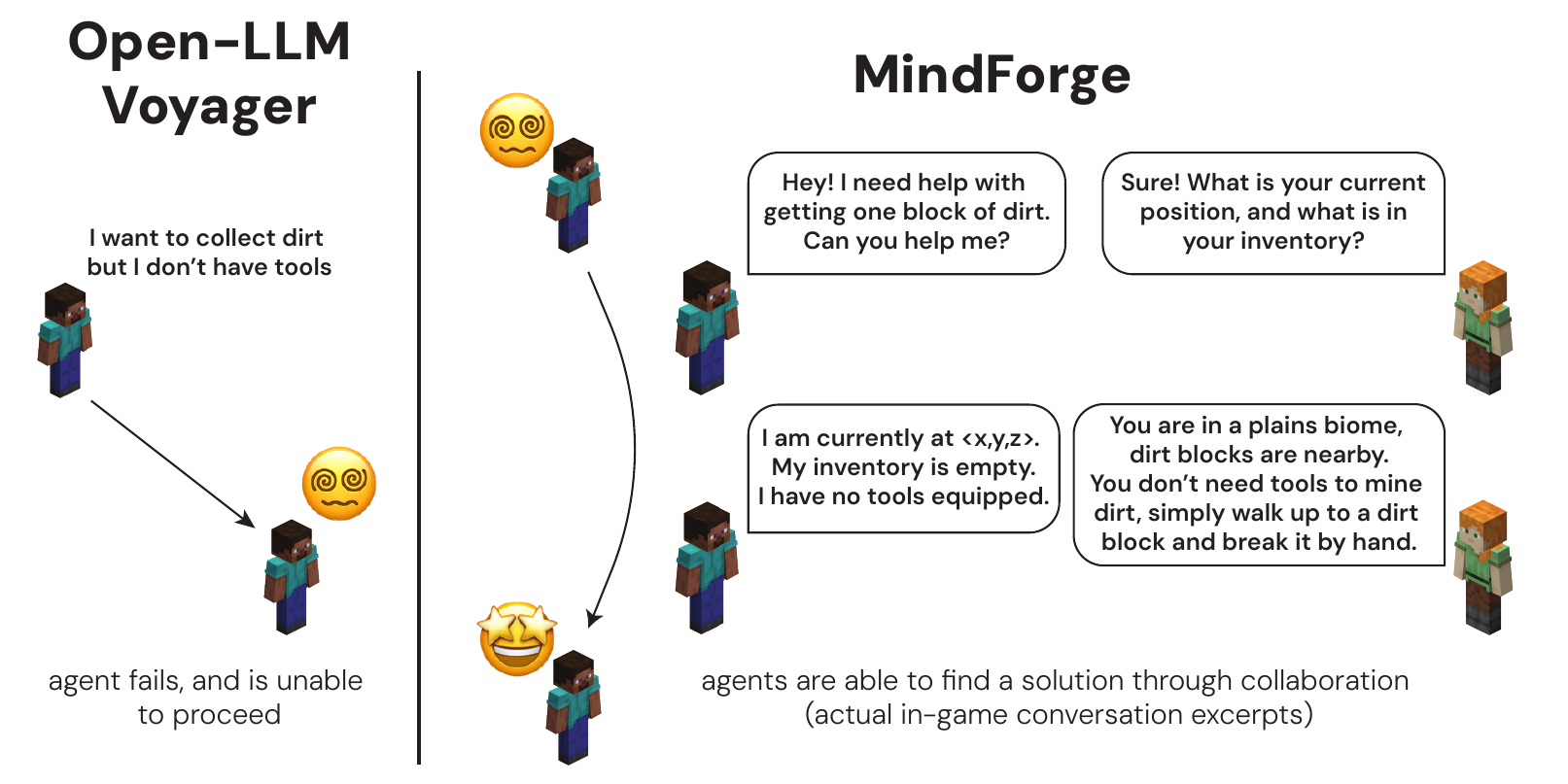}
         \caption{}\label{sfig:teaser}
     \end{subfigure}
     \hfill
     \begin{subfigure}[b]{0.5\textwidth}
         \centering
         \includegraphics[width=\linewidth]{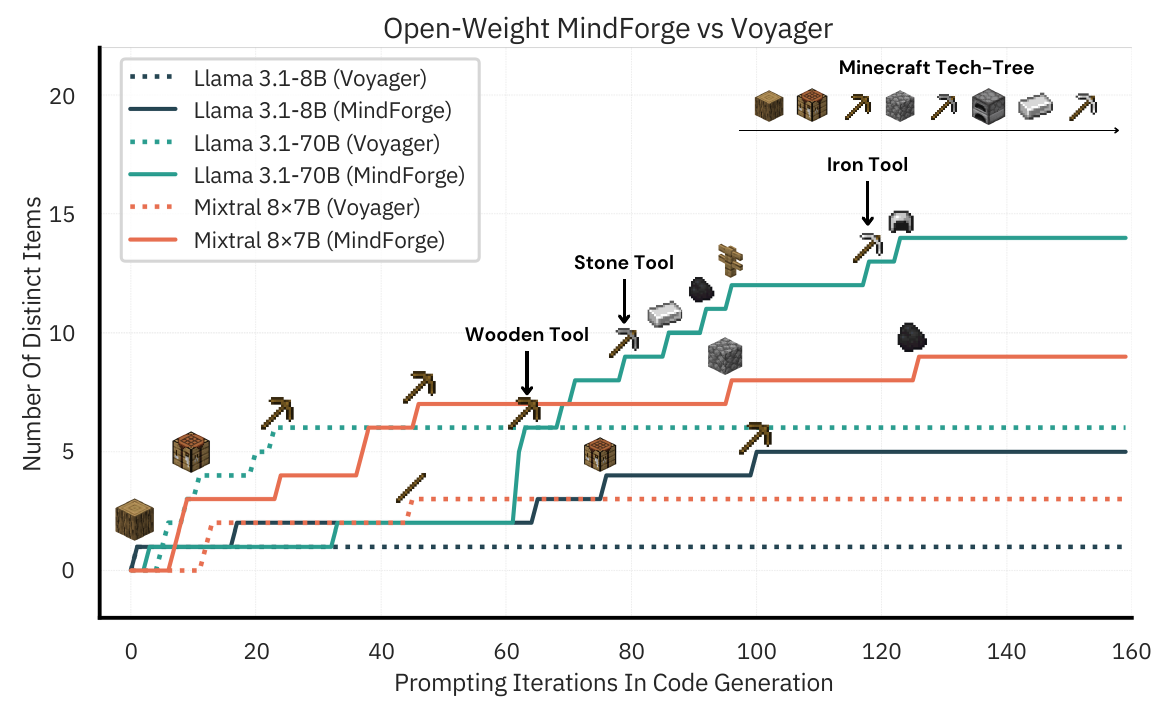}
         \caption{}\label{sfig:mind_v_voy}
     \end{subfigure}
     \caption{\collabvoyager enables agents to (a) collaborate through structured theory-of-mind representations, which (b) leads to improved performance compared to Voyager when powered by open-weight LLMs, averaged across 3 runs.}
\end{figure*}

\section{Introduction}\label{sec:intro}
The development of generally capable agents marks a significant shift in advancing artificial intelligence, transitioning from assimilating data to generating novel knowledge through embodied interactions with open-ended environments \citep{kolve2017ai2, savva2019habitat, puig2018virtualhome, shridhar2020alfred}. Classical approaches leveraging reinforcement learning \citep{schulman2017proximal, hafner2023mastering} and imitation learning \citep{zare2024survey} often struggle with generalization and exploration, as agents tend to converge on repetitive behaviors in static environments \citep{cobbe2019quantifying}. To address these limitations, researchers have sought to emulate human-like lifelong learning capabilities, developing systems that can continuously acquire, update, and transfer knowledge over extended periods \citep{parisi2019continual, wang2023comprehensive}.The advent of large language models (LLMs) has accelerated this pursuit, enabling the development of agents such as Voyager \citep{wang2023voyager} that can apply internet-scale knowledge to continuously explore, plan, and acquire new skills in partially observable, open-ended environments such as Minecraft. 

Despite their promise, we argue that state-of-the-art lifelong learning agents like Voyager face a crucial limitation: they learn in isolation, neglecting a fundamental aspect of human intelligence\textemdash the social context.  So central is the social context to our existence that the Social Intelligence Hypothesis posits that our cognitive capabilities evolved primarily to navigate the complexities of social life \citep{humphrey1976social, dunbar1998social}. This isolated learning becomes particularly problematic when coupled with these agents' reliance on closed LLMs like GPT-4. \citet{wang2023voyager} note that \textit{``VOYAGER requires the quantum leap in code generation quality from GPT-4 which GPT-3.5 and open-source LLMs cannot provide''}. Our experiments (\Cref{sec:experiments}) corroborate this observation, demonstrating that replacing GPT-4 with open LLMs renders the Voyager agent incapable of even the most rudimentary tasks, such as collecting dirt in Minecraft. We ask: can an open LLM-powered agent, which struggles with basic tasks, improve through collaboration with others?

To address this question, we introduce the \collabvoyager framework, inspired by cultural learning, which contends that individuals internalize skills and strategies by observing others' perspectives, using Theory of Mind (ToM) to infer goals, beliefs, and intentions and thereby deepen social learning \citep{tomasello1993cultural, tomasello2016cultural, premack1978does, baron1999evolution, devaine2014social}.
Our framework reframes \textit{knowledge distillation}~\citep{hinton2015distilling} as a social, test-time process: agents refine their beliefs through multi-turn dialogue with expert peers, without gradient updates. This interactive distillation supports \textit{test-time compute scaling} \citep{snell2024scaling}, enabling agents to improve performance via communication, memory retrieval and perspective-taking. By structuring communication across turns, \collabvoyager also contributes to \textit{multi-turn LLM agent} research~\citep{zhang2023building, du2023improving}, showing that open models can match GPT-4 through collaboration alone. Moreover, our analysis of weak-weak settings highlights a compute-performance tradeoff at inference, linking collaboration to test-time compute considerations~\citep{ snell2024scaling}.

\Cref{sfig:teaser} provides an overview of our agents in action. Concretely, \collabvoyager extends Voyager by incorporating three key improvements:  (1) \textbf{perspective-taking} through a causal ToM template \cite{gandhi2024understanding} linking agent percepts, beliefs, desires, and actions; (2)  a \textbf{natural language communication interface} between agents; and (3) memory updates following the Soar cognitive architecture \cite{laird2019soar}, including \textbf{semantic memory} of task and environment-related knowledge and \textbf{episodic memory}.

Empirically, we instantiate \collabvoyager agents in \textit{Instructed Learning} (mixed-expertise) and \textit{Collaborative Learning} (matched-expertise) settings within the cultural learning framework \citep{tomasello2016cultural}. \collabvoyager agents powered by open-weight LLMs outperform the original Voyager architecture \citep{wang2023voyager} on two seemingly trivial tasks by more than $40 \%$. We identify Voyager’s specific failure cases: (1) \textbf{false beliefs} and (2) \textbf{code generation}, and show how our method overcomes them through communication and perspective-taking. To assess lifelong learning, we run \collabvoyager on the Minecraft tech-tree and find that it surpasses open-weight Voyager variants in both key tech-tree milestones and the number of unique items obtained. ~\Cref{sec:colab-learning} further shows that when both agents begin with the same limited capabilities, a single round of GPT-4 seeding followed by just seven turns of peer-to-peer dialogue lifts the population-level success rate from $62 \%$ to $79 \%$, revealing a Condorcet-style ``many-minds'' boost that stems from richer interaction rather than larger models. This generalization in post-collaborative settings paves the way for agents that can continuously adapt through social interaction, reducing the reliance on large-scale computing resources for training and fine-tuning the underlying language models.

\section{Related Work}\label{sec:related-work}


\paragraph{Theory of Mind in Embodied Agents} Early attempts to equip machines with mental-state reasoning drew on deep learning~\citep{rabinowitz2018machine, chuang2020using, shu2021agent}, reinforcement learning~\citep{grassiotto2021cogtom}, and Bayesian models~\citep{baker2011bayesian}. These systems were either confined to toy 2-D worlds~\citep{rabinowitz2018machine, chuang2020using} or failed to generalize to richer environments~\citep{shu2021agent}. The emergence of large language models (LLMs) reignited interest in ToM, sparking evaluations of their putative capabilities~\citep{kosinski2023theory, bubeck2023sparks, ullman2023large, gandhi2024understanding}. Although evidence so far suggests that LLMs lack genuine ToM, their breadth of knowledge has inspired a new wave of prompt-based frameworks~\citep{wilf2023think, zhou2023far, sclar2023minding, hao2023reasoning}: the first two focus on perspective-taking, while the latter pair ground reasoning in explicit causal models. BIP-ALM remains the sole example that integrates such a framework with an LLM~\citep{jin2024mmtom, li2023theory}.

\paragraph{LLM-based Multi-Agent Systems} LLMs have demonstrated overwhelming success in performing a variety of single-agent tasks ranging from question answering and text generation to code synthesis and complex reasoning in embodied settings \citep{wang2023voyager, ahn2022can, lin2023text2motion}. This success has started a growing interest in developing LLM-based multi-agent systems \citep{zhang2024combo, zhang2023building, du2023improving, talebirad2023multi, vezhnevets2023generative}, where language-based communication is a central component. For instance, CoELA \cite{zhang2023building} introduce a modular multi-agent framework that integrates memory, perception and execution, while COMBO \cite{zhang2024combo} present a multi-agent cooperative framework, that deals with collaboration in partial egocentric views of the environment. However, these frameworks lack an explicit belief representation and Theory-of-Mind reasoning. Recently, benchmarks like MineCollab \cite{white2025collaborating}, designed for collaborative reasoning, find that current LLM agents are not suited for multi-agent collaboration in long-horizon embodied scenarios, a gap that \collabvoyager addresses by endowing agents with structured perspective taking capabilities. 

\paragraph{Single-agent Frameworks}
Recent research on long-horizon tasks in single-agent systems differs architecturally from \collabvoyager's unified collaborative learning framework. For example, ExpeL~\cite{zhao2024expel} and CLIN~\cite{majumder2023clin} emphasize experiential learning through vectorized trajectories or causal abstractions but lack explicit belief graphs and integration across episodic, semantic, and procedural memories. Frameworks like DEPS~\cite{wang2023describe} and ADaPT~\cite{prasad2023adapt} enable in-episode self-correction via iterative refinement, yet insights are often ephemeral, confined to the context window without persistent storage or structured models of others' beliefs. SSO~\cite{nottingham2024skill} provides procedural memory through skill libraries but omits episodic/semantic layers and social reasoning for targeted teaching or belief revision. Similarly, Adam~\cite{yu2024adam} structures world knowledge as causal graphs in open environments, while Optimus-1~\cite{li2024optimus} builds hierarchical semantic graphs from experiences.

\section{Background}\label{sec:bg}

\paragraph{Voyager} Voyager agents consist of three main components: (1) an automatic curriculum generator, (2) a skill library, and (3) an iterative prompting mechanism. The automatic curriculum proposes progressively complex tasks, driving continuous learning. The skill library stores and retrieves learned behaviors, facilitating knowledge accumulation and transfer. The iterative prompting mechanism generates and refines executable code based on environmental feedback, allowing for adaptive behavior. Voyager demonstrated significant capabilities in item discovery, technology progression, and skill generalization. This system established a new benchmark in open-world AI agents, showcasing the potential for lifelong learning and adaptation in complex, dynamic environments.  The Voyager framework serves as a base for our lifelong \textit{collaborative} learning agents within Minecraft. 
 
\paragraph{BigToM Causal Template} 
The BigToM causal template \citep{gandhi2024understanding}, provides a structured framework for representing Theory of Mind (ToM) scenarios as causal graphs. This template comprises key components including context, desire, percept, belief, causal event, and resulting actions. By representing ToM scenarios in this causal structure, the template allows for systematic manipulation of variables such as an agent's perceptions, beliefs, and actions in response to environmental changes. This approach enables the creation of diverse, well-controlled ToM scenarios, capturing the complex causal relationships between an agent's mental states and their behavior in social situations. The causal graph representation facilitates a more nuanced and systematic exploration of ToM reasoning, allowing for the examination of various inference patterns such as forward belief, forward action, and backward belief inferences in social cognitive tasks.

\begin{figure*}[t]
    \captionsetup[subfigure]{belowskip=-3pt}
     \centering
     \begin{subfigure}[b]{0.55\textwidth}
         \centering
         \includegraphics[width=\columnwidth, trim={1.5cm 6.85cm 3cm 5.7cm},clip]{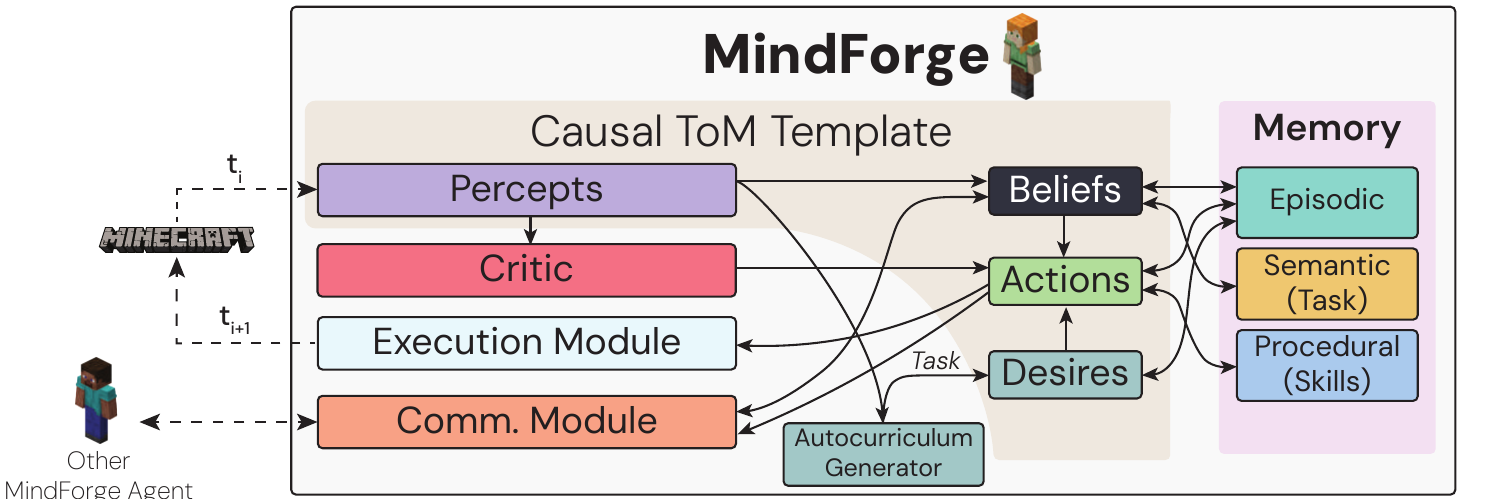}
         \caption{\collabvoyager Architecture}\label{sfig:mindcraft-arch}
     \end{subfigure}
     \hfill
     \begin{subfigure}[b]{0.43\textwidth}
         \centering
         \includegraphics[width=\columnwidth, trim={3cm 6.7cm 4.5cm 6.3cm},clip]{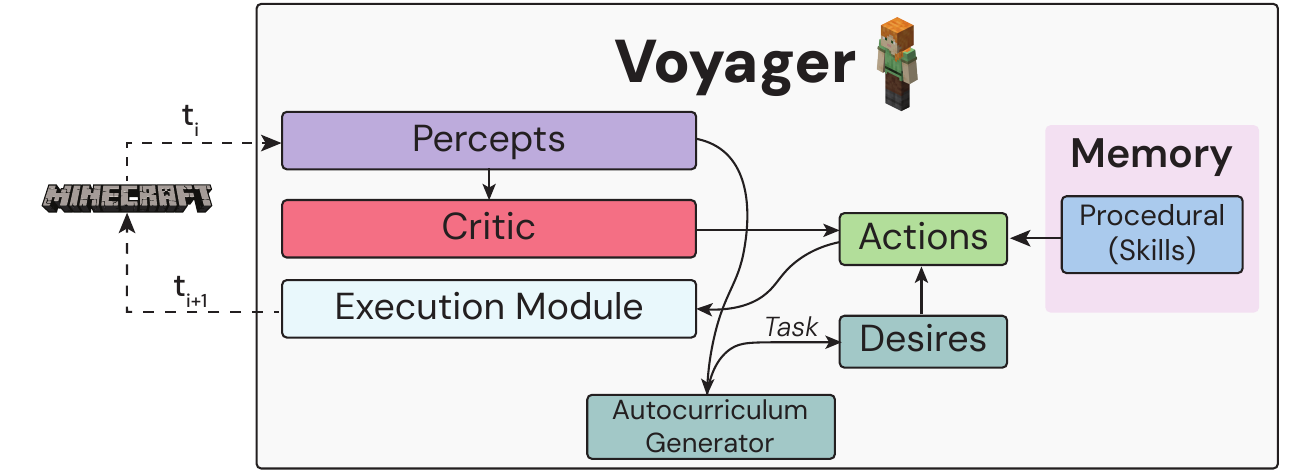}
         \caption{Voyager Architecture}\label{sfig:voyager-arch}
     \end{subfigure}
     \caption{Cognitive architectures of \collabvoyager and Voyager respectively. The Voyager architecture has been re-framed within the scope of our cognitive architecture. \collabvoyager expands the original Voyager framework to enable communication through theory of mind together with additional memory subsystems to enhance the lifelong learning capabilities of the agent.}
     \label{fig:mindcraft-voyager}
\end{figure*}

\section{Method}
\label{sec:methodology}

To grant \collabvoyager agents the ability to learn through collaborative social interactions, we introduce the following components for communication and perspective-taking:
(i) an \textbf{explicit structured representation} linking percepts, beliefs, desires, and actions (\Cref{ssec:world-model}), 
(ii) \textbf{perspective taking}  so that the agent can reason over self-beliefs and those about its collaborators   (\Cref{ssec:soc-aware-persp}), 
(iii)
a \textbf{communication medium} that the agents can use to share information in natural language (\Cref{ssec:comm-medium}),
and (iv) the ability to utilize different types of \textbf{memory subsystems} (\Cref{ssec:memory-subsys}).

\vspace{-0.2cm}

\subsection{Structured Representation of Own Mental States: Beliefs, Desires, and Intentions}\label{ssec:world-model}
For \collabvoyager agents to develop effective theory of mind capabilities, they must first possess structured representations of their own mental states. 
This explicit modeling of the self is crucial for \textit{recursive social reasoning}: enabling agents not only to reason about their own beliefs, goals, perceptions, and how these might lead to their actions, but also to form a similar mental model for other agents. 
This mirrors findings in developmental psychology, where children first develop an awareness of their own thoughts, feelings, and goals before being able to infer the mental states of others \citep{Wellman1990-qh, Perner-book}. This is further supported by neuroscientific research that has also identified distinct neural mechanisms for self-cognition and social cognition \citep{Ochsner2004-jd, Saxe2003-zr}. 

We draw on the classical Belief-Desire-Intention (BDI) framework \citep{rao1995bdi} and operationalize it using the BigToM causal template (\Cref{sec:bg}) \citep{gandhi2024understanding}. In \collabvoyager, \textit{beliefs} represent the agent's current understanding of the world inferred from sensory perception, including interactions with other agents. \textit{Desires} encode the agent's goals, both generated by an automatic curriculum and those arising from collaborative interactions. \textit{Intentions} correspond to concrete executable actions, implemented as generated code\textemdash and so, consistent with BigToM \citep{gandhi2024understanding}, we treat ``intentions'' and ``actions'' interchangeably in implementation. However, we note that this equivalence does not always hold conceptually: intentions refer to internal commitments or plans, while actions are observable behaviors that may diverge from intentions. For a deeper treatment of this distinction, see \citep{hung2024discontent}.

 
\paragraph{Belief Structure.} We categorize a \collabvoyager agent's beliefs into four key categories, illustrated in \Cref{fig:beliefs}: i. \textit{perception-related beliefs}, formed from direct sensory input via the Minecraft API; ii. \textit{task-related beliefs}, inferred from the agent's reflection on its upcoming objectives; iii. \textit{interaction-related beliefs}, derived from chat messages during task execution; and iv. \textit{partner-related beliefs}, which track others' mental states. Each belief type is generated via a separate LLM call, with prompts conditioned on current percepts, memory retrievals, and conversation history (see \Cref{appendix:prompts}). This modular belief formation feeds into the causal ToM template \Cref{sfig:mindcraft-arch}, which governs downstream action generation. Structured self-modeling thus serves as the foundation for cultural learning and perspective-taking. 


\subsection{Theory of Mind: Modeling Partner Perspectives}\label{ssec:soc-aware-persp}

\begin{figure}[t!]
\centering
\includegraphics[width=0.65\columnwidth, trim={5mm 5.2cm 2cm 0.9cm},clip]{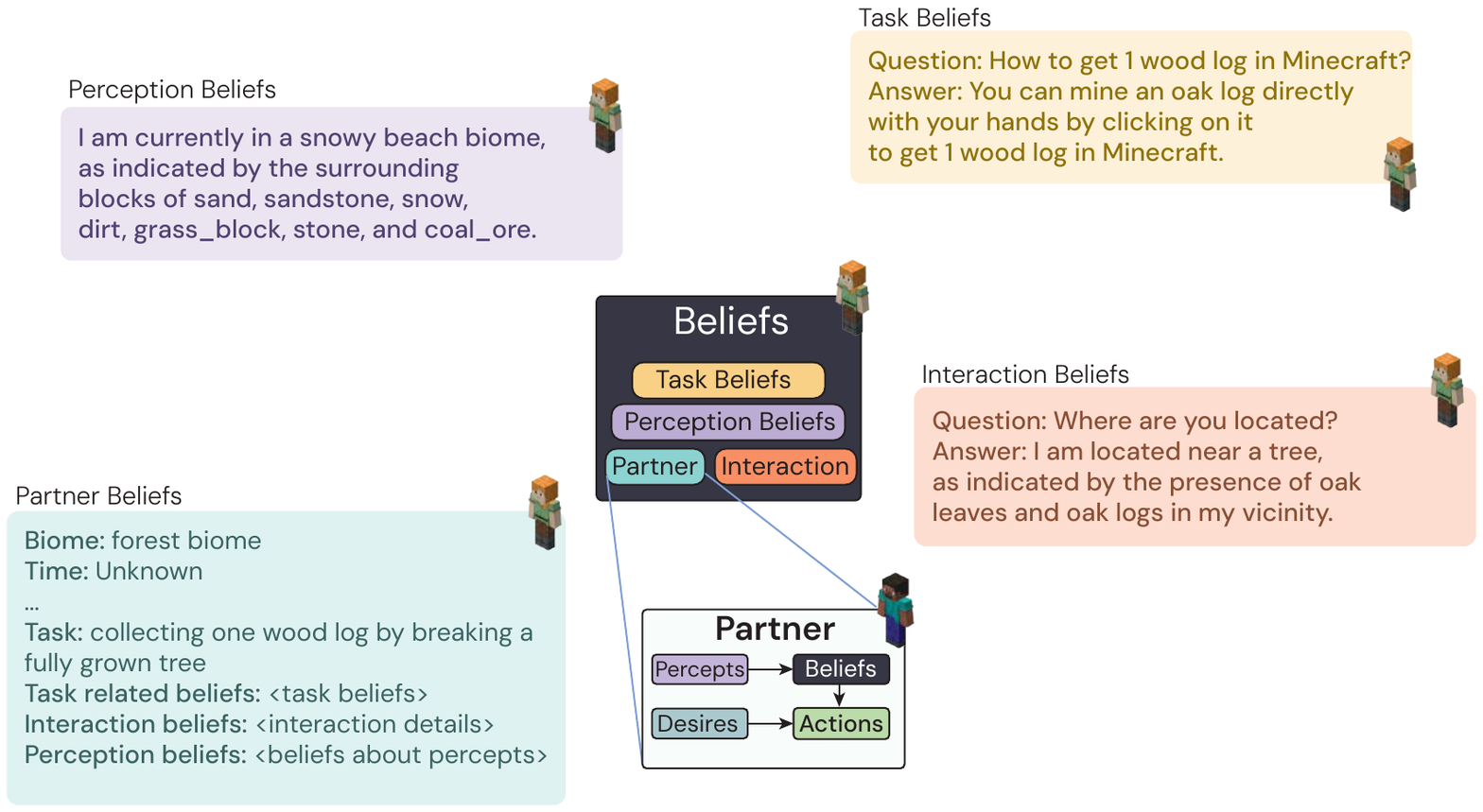}
\caption{Each agent maintains a set of internal beliefs that it can exploit as needed. In order to enable ToM capabilities agents also maintain a belief over the beliefs, actions and desires of their collaborative counterparts.}
\label{fig:beliefs}
\end{figure}

Building on the structured representation of their own mental states, \collabvoyager extend this modeling to others via Theory of Mind\textemdash the capacity to attribute beliefs, desires, and intentions to others. 
Each \collabvoyager agent achieves this by maintaining a partner-specific BigToM graph within its own beliefs, allowing it to reason about the internal state of each conversational partner (see \Cref{fig:beliefs}). These representations guide how an agent interprets and responds to incoming communication. 

During collaboration, an agent uses its internal model of its partner to simulate what the partner knows, believes, or needs. The underlying LLM receives this partner model along with the conversation history to generate responses incorporating the collaborator's point of view. As interaction proceeds, the partner beliefs are continuously updated using new  received through the communication channel (\Cref{fig:beliefs-evolution-rounds}). 
This dynamic modeling of partner mental states endows \collabvoyager agents with one of the most crucial abilities within the cultural learning framework: reasoning about partner perspectives \citep{tomasello1993cultural, tomasello2016cultural}. For instance, an expert agent can diagnose a novice's misconception and tailor its feedback accordingly, while a novice can ask more targeted questions with fewer digressions by simulating what the expert knows. See \Cref{appendix:perspective} for more details on how perspective-taking is integrated into the communication module, and in-game examples.

\subsection{Communication Medium}\label{ssec:comm-medium}

To take advantage of representations of each other's mental models and collaborate effectively, agents need to be able to communicate with each other. We operationalize communication in natural language, integrating the \collabvoyager framework's communication system with Minecraft's native in-game chat functionality. 
Specifically, once a \collabvoyager agent fails to complete a task, it initiates a communication round with its partner where it leverages its BigToM representation to ask for help. This system enables agents to exchange textual information and, crucially, to integrate feedback from collaborators directly into their belief systems. Concretely, we utilize the Mineflayer API \citep{mineflayer} to extract and process chat data from the Minecraft environment. By leveraging the game's existing chat system, the framework ensures that collaborative interactions occur in a context-appropriate manner, closely mirroring human-to-human communication within Minecraft. 

To handle the resulting conversation, we use a dedicated LLM instance within the \collabvoyager agent which is tasked with synthesizing and interpreting the conversation into the structured BigTOM BDI template, generating both interaction beliefs and partner beliefs (see prompts in \Cref{appendix:prompts}). This instance does not directly influence any other parts of the agent.

\subsection{Memory Subsystems}\label{ssec:memory-subsys}

A \collabvoyager agent's memory consists of three subsystems drawing on research about human memory: (i) \textbf{episodic}, (ii) \textbf{semantic}, and (iii) \textbf{procedural} memory.

\textbf{Episodic memory} deals with pertinent information about the agent's past attempts at a task and social interactions with other agents. 
Recollections of past episodes serve as context for humans to interpret others' behaviors, leading to divergent future actions in the same situation in response to different recollections \citep{bornstein2017reminders}. 
Humans also tend to summarize information from their social interactions, but at the same time are capable of preferentially accessing episodes when the credibility of certain summaries is low due to the lack of evidence \citep{Klein2009-vz}. Moreover, findings also suggest that there are processes in the brain for updating these summaries as new episodes transpire \cite{Klein1993-vu,Klein2009-vz}. Inspired by complementary learning systems theory, Larimar \citep{das2024larimar} enhances LLMs with a distributed episodic memory module, enabling dynamic knowledge updates without any gradient-based learning. Furthermore, Optimus-1 integrates past experiences in a multimodal memory storage mechanism and abstracted experience pool used for summarization of past experiences. For a \collabvoyager agent, the role of episodic memory is twofold. Firstly, it stores details about its partners revealed during previous conversations that can be exploited for perspective-taking in future communication. Secondly, we store past actions that led to failure and retrieve them based on their similarity with the task at hand. Thus, a \collabvoyager agent can track both the progress of its partner and learn from past failures.

\textbf{Semantic memory} constitutes an individual's store of knowledge about the world. The content of semantic memory is abstracted from the experience and is therefore said to hold summarized generalizations, likely without references to specific experiences \cite{Binder2011-gq, Klein2009-vz, Klein1993-vu}. For a \collabvoyager agent, the semantic memory stores information related to the task and environment. It stores information relevant to completing a task; for instance, the prerequisites for being able to mine dirt in the game world. Such a piece of information is not necessarily tied to one specific episode. Rather, it is generally useful information about the world and can be distilled from the episodic memory.

Finally, \textbf{procedural memory} stores the skills that the agent learns and is analogous to the skill library present in the original Voyager framework \cite{wang2023voyager}.  The skills stored in the procedural memory are reusable pieces of code that the agent can use for solving problems, allowing the agent to take on more complex tasks while avoiding re-learning elements previously used successfully. 
As illustrated in \Cref{sfig:mindcraft-arch}, episodic and semantic memory stores influence the agent's beliefs, while procedural memory provides the routines that inform the agent's next action. Additional information about the memory subsystems can be found in \Cref{appendix:memory}, where we provide implementation details, and an ablation for the episodic memory component, where we show that omitting it from \collabvoyager leads to a drop in success rate. 

\section{Experiments}
\label{sec:experiments}

\subsection{Experimental Setup}
\label{ssec:setup}

Our \collabvoyager framework introduces three advances that let agents improve continuously through social exchange: (1) explicit theory-of-mind representations for reasoning about peers, (2) Natural-language dialogue for coordination, and (3) integrated semantic and episodic memory to retain experience.
We evaluate these additions in the same Minecraft world used by Voyager~\citep{wang2023voyager}, built on MineDojo~\citep{fan2022minedojo} with motor control via Mineflayer~\citep{mineflayer}. Component LLMs communicate through DSPy~\citep{khattab2024dspy, khattab2022demonstrate} and LangChain~\citep{langchain2024} for generation and embedding. In the base experimental setting, we follow Voyager \cite{wang2023voyager} and allow the agent to execute 4 sequential actions to solve a task. \collabvoyager agents that interact (see \Cref{ssec:collab}) interleave one communication round (via the in-game chat) between two environmental actions. A snapshot of this setup can be seen in~\Cref{fig:code-belief-correction}.
For isolated tasks we measure the fraction of successes over $24$ independent trials (one task per trial). In tech-tree experiments we follow Voyager's protocol, tracking milestone achievements and unique items collected. To ensure comparability, we preserve Voyager's original prompt structure and workflow throughout.

\subsection{Voyager With Open-Weight LLMs}
\label{ssec:baselines}

To establish non-interactive performance baselines, we replace GPT-4 \citep{achiam2023gpt} with three open-weight LLMs in the Voyager framework \citep{wang2023voyager}: Mistral-7B \citep{jiang2023mistral7b}, Mixtral-8x7B \citep{jiang2024mixtral}, and Llama 3.1-8B \citep{dubey2024llama}, selected for their long context windows and manageable size. Testing focuses on two fundamental Minecraft tasks: collecting single blocks of dirt and wood.
Results in \Cref{table:voyager-base} reveal that while GPT-4 achieves perfect performance, open-weight LLMs struggle significantly even with these basic tasks without collaboration. This aligns with \citet{wang2023voyager}'s observation that GPT-4's superior code generation capabilities enable it to obtain $5.7\times$ more unique items than other models, establishing a clear baseline for measuring the impact of our collaborative approach.

\begin{table}[hbt]
    \scriptsize
  \caption{Voyager \cite{wang2023voyager} powered by open-weight LLMs (even fine-tuned for Minecraft) struggles with basic tasks. Closed-source models like GPT-4 exhibit perfect accuracy. A single \collabvoyager agent bridges the performance gap between Mistral-7B and GPT-4 on both wood and dirt collection, showing significant gains compared to other learning techniques such as PEFT.}
    {\tabcolsep=9pt\def\arraystretch{1.1}
    \begin{tabularx}{\columnwidth}{@{}>{\arraybackslash}l *6{X}}
      \toprule
      \textbf{Task} & \textbf{Mistral-7B} & \textbf{Mixtral-8x7B} &\textbf{ Llama 3.1-8B} & \textbf{Mistral-7B (PEFT)} & \textbf{Mistral-7B (MindForge)} & \textbf{GPT-4} \\ 
      \midrule
      Dirt & $7\%$ & $27\%$ & $4\%$ & $7\%$ & $37.5\%$ & $100\%$ \\
      Wood & $4\%$ & $27\%$ & $7\%$ & $12.5\%$ &  $33.3\%$ &$100\%$ \\
      \bottomrule
    \end{tabularx}}
    \label{table:voyager-base}
\end{table}

\begin{figure}[t] 
    \centering 

    \begin{subfigure}[t]{0.58\textwidth} 
        \centering
        \includegraphics[width=\linewidth,trim={3.35cm 6.2cm 2.1cm 5.1cm},clip]{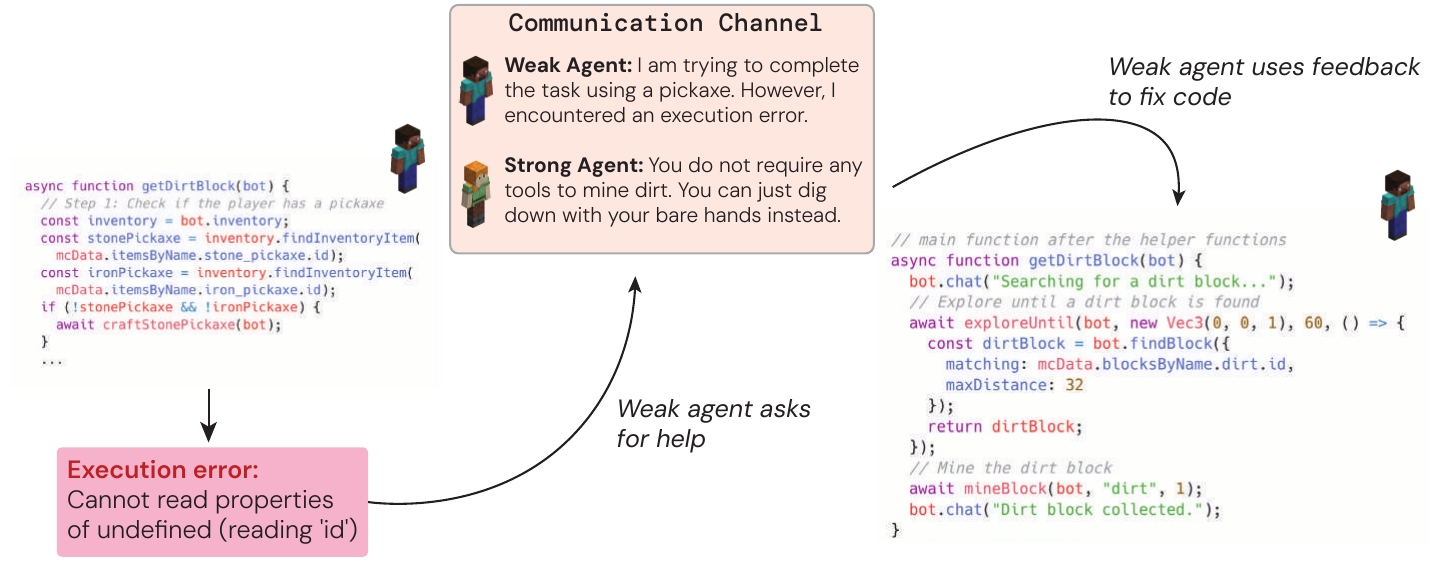} 
        \caption{} \label{fig:code-belief-correction}
    \end{subfigure}
    \hfill 
    \begin{subfigure}[t]{0.40\textwidth} 
        \centering
        \includegraphics[width=\linewidth,trim={1.05cm 7.3cm 1.25cm 8.5cm},clip]{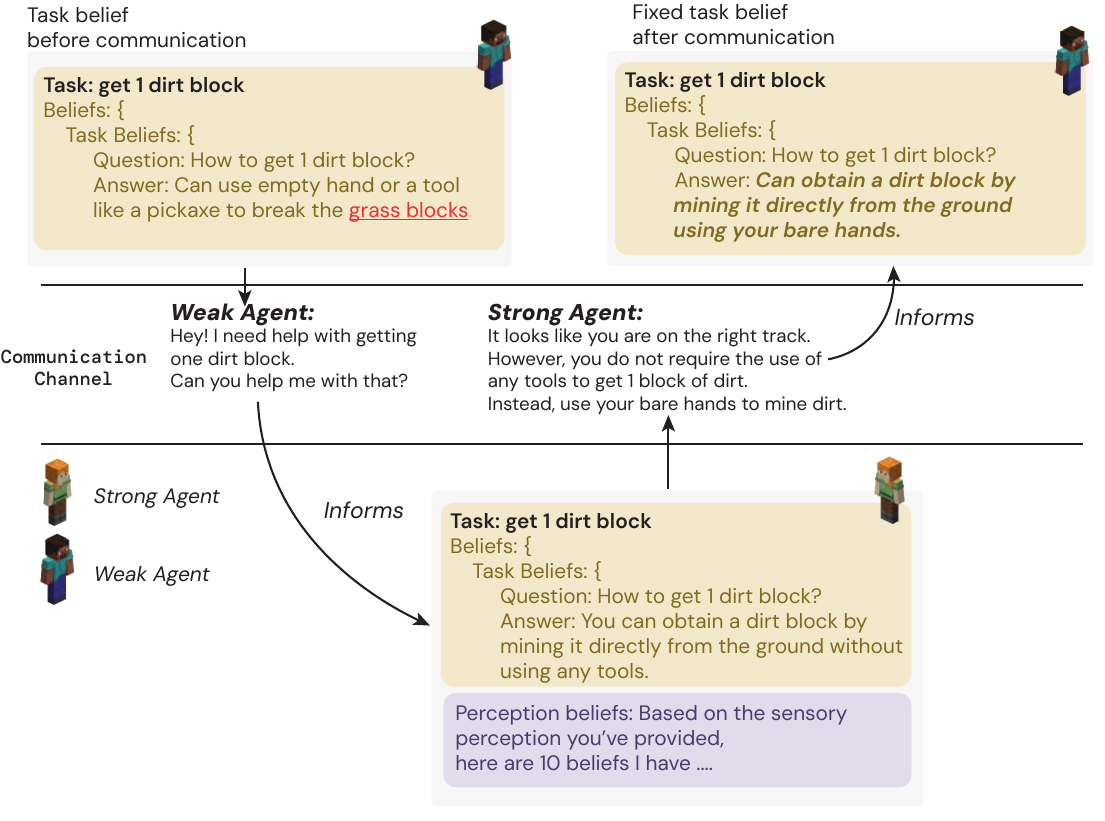} 
        \caption{} \label{fig:comm-internal-improvement}
    \end{subfigure}

    \caption{Failure mode correction. (a) Communication with an expert agent fixes the code error. (b) A weaker agent's incorrect task-related belief (initially believing it is mining `grass blocks' and needs tools) is corrected by a stronger agent through communication (e.g., stating `dirt blocks' do not require tools), leading to an updated task understanding.} 
    \label{fig:combined-belief-and-comm} 
\end{figure}

\paragraph{Primary Failure Cases} While \citet{wang2023voyager} only discuss code-generation failures, we find two prominent failure cases for the open-weight Voyager setup: (1) \textbf{false beliefs} and (2) \textbf{faulty code generation}. Importantly, we discover that the critic-based self-improvement component in Voyager fails to recover from initial mistakes and subsequently fails to solve the task. As shown in \Cref{fig:comm-internal-improvement}, a Mistral-7B powered Voyager fails at dirt collection due to incorrect assumptions about tool requirements and lack of basic knowledge pertaining to Minecraft. Similarly, \Cref{fig:code-belief-correction} showcases how feedback from a stronger agent can solve code misconceptions about the Mineflayer API \citep{mineflayer}.

\paragraph{Fine-Tuning Does Not Help}
Parameter-efficient tuning often helps LLMs~\citep{han2024parameter}, yet adapting Mistral-7B with GPT-4 execution traces (reasoning + code) and Minecraft manuals (domain knowledge;~\Cref{appendix:fine}) produced no benefit. In cultural-learning terms~\citep{tomasello1993cultural, tomasello2016cultural} this is imitation learning, but the tuned model still scores $7\%$ on dirt collection ($\pm0$) and $12.5\%$ on wood collection ($+8.5$) as shown in ~\Cref{table:voyager-base}. Post-hoc analysis shows the tuning makes outputs wordier, blurs decision-making, and erodes instruction-following, indicating that naively imitating unstructured traces is counter-productive.

\subsection{Introducing Collaboration: \collabvoyager Agents}
\label{ssec:collab}
Having shown that PEFT-based imitation learning fails to improve agent capabilities, we evaluate our proposed \collabvoyager framework. We adopt an instructive learning approach within the cultural learning framework \citep{tomasello2016cultural}, where an expert agent guides a novice through communication and perspective-taking. This multi-agent collaboration effectively bridges the performance gap between open-weight LLMs and GPT-4, addressing the failure cases identified in \Cref{fig:combined-belief-and-comm}.

\begin{figure}[t]
\centering
\begin{minipage}{0.58\textwidth}
\captionof{table}{Fraction of successful \collabvoyager agents across $24$ individual trials; dirt and wood collection. Standard Voyager \cite{wang2023voyager} setup: $4$ attempts per task, with a communication round interleaved where appropriate. See also~\Cref{fig:comm-rounds} and~\Cref{fig:weak-weak}.}
  \scriptsize
  {\tabcolsep=5pt\def\arraystretch{1.1}
  \begin{tabularx}{\textwidth}{@{}>{\arraybackslash}l *4{X}}
  \toprule
  \bf{Setting} & \textbf{Task} & \textbf{Mistral-7B} & \textbf{Mixtral-8x7B} & \textbf{Llama 3.1-8B} \\ 
  \midrule
    w/o interactions & Dirt & $37.5$\% & $29.15$\% & $12.5\%$ \\
    instructive setting, human expert & Dirt & $+45.83$  & $+57.85$ & $+45.83$ \\
    instructive setting, GPT-4 expert & Dirt & \bm{$+16.16$} & \bm{$+37.51$} & \bm{$+4.1$}\\
    instructive setting, Llama 3.1-70B & Dirt & $+8.33\%$ & - & - \\
  collaborative setting, no expert & Dirt & $-4.2$ & $+0$ & $-8.5$ \\
  \midrule
  w/o interactions & Wood & $33.3$\% & $50\%$ & $7\%$ \\
  instructive setting, GPT-4 expert & Wood & \bm{$+12.52$} & \bm{$+20.8$} & \bm{$+5.5$} \\
  instructive setting, Llama 3.1-70B & Wood & $+8.36\%$ & - & - \\
  collaborative setting, no expert & Wood & $-12.5$ & $-4.15$ & $+0$ \\
  \bottomrule
  \end{tabularx}}
  \label{table:main}
  \end{minipage}
  \hfill
  \begin{minipage}{0.40\textwidth}
       \centering
         \includegraphics[width=\linewidth,trim={0cm 0cm 0cm 0cm},clip]{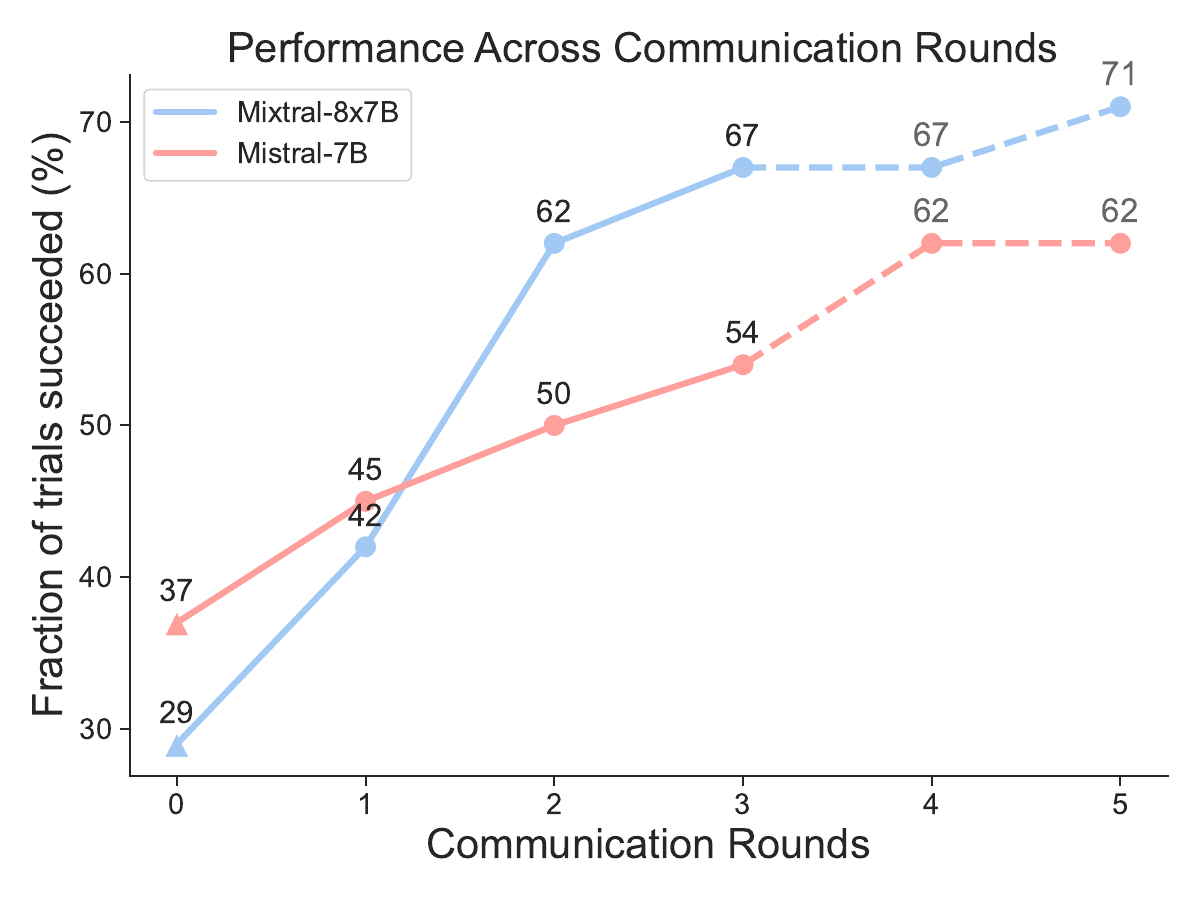}
         \caption{\textbf{Instructive Setting (GPT-4); Dirt Collection.} Performance scales with communication rounds. \collabvoyager w/o interactions as a baseline. Actions are interleaved with communication rounds. Dashed\quad(– –): extra collaboration rounds. $\triangle$ points: cold-start (4 actions) without interactions.}\label{fig:comm-rounds}
  \end{minipage}
\vspace{-5pt}
\end{figure}

\paragraph{Feasibility: Human (Expert) Collaboration Improves Performance} To establish feasibility, we first conduct a wizard-of-oz study where a human expert interacts with an open-weight LLM-powered \collabvoyager agent through the in-game chat. As shown in \Cref{table:main}, human expert collaboration achieves on average a $+57.85\%$ increase in task completion rate for dirt collection over the non-interactive baselines. This demonstrates both the viability and flexibility of our framework in enabling productive human-agent interactions for collaborative tasks.


\paragraph{GPT-4 (Expert) Collaboration Maintains Performance Benefits} We replace the human expert with a GPT-4-powered \collabvoyager agent that collaborates with an open-weight LLM agent. The expert agent infers its partner's failure cases through perspective-taking based on natural language communication in the game chat. Unlike human collaboration, agents cannot directly access each other's internal beliefs and must rely on Theory of Mind through communication to model their counterpart's mental state. This perspective-taking enables the expert to accurately diagnose and address the weaker agent's misconceptions. Notably, this setup generalizes the classic notion of knowledge distillation beyond logits or model weights: here, distillation unfolds through social interaction, where the expert agent transmits structural knowledge, corrects false beliefs, and shapes the student’s internal representations in real time. As shown in~\Cref{table:main}, this approach significantly improves performance compared to the non-interactive baselines, achieving on average a $16.1\%$ increase in task completion rates for dirt collection and a $13\%$ increase for wood collection tasks respectively. A sample conversation can be found in \Cref{appendix:sample-conv}.

\paragraph{Ablations} To elucidate the \collabvoyager framework's components, we conducted targeted ablations. The perspective-taking module, enabling agents to reason over partners' beliefs, boosts task completion by up to 13\%, as evidenced by comparisons across multiple communication rounds (see \Cref{tab:perf-perspective}, \Cref{appendix:perspective}). Our structured ToM representation outperforms its unstructured variant in complex Minecraft tasks (\Cref{tab:perf-perspective-unstructured}). Additionally, ablating the episodic memory component reveals that retaining past failures enhances performance (\Cref{table:episodic-ablation}, \Cref{appendix:memory}).

\begin{figure*}[t]
\centering

\begin{minipage}[t]{0.58\textwidth}
    \centering
    \includegraphics[width=\columnwidth, trim={3.5cm 7.6cm 7.3cm 7.0cm}, clip]{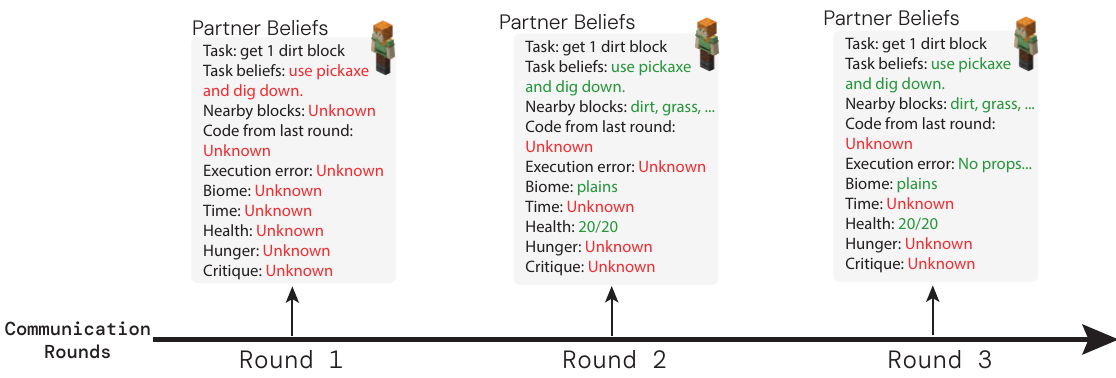}
    \captionof{figure}{Partner beliefs of the strong agent progressively improve as the number of communication rounds increase. Subsequently, these beliefs are used to reason throughout the conversation and take perspective.}
    \label{fig:beliefs-evolution-rounds}
\end{minipage}
\hfill
\begin{minipage}[t]{0.40\textwidth}
    \centering
    \includegraphics[width=\columnwidth]{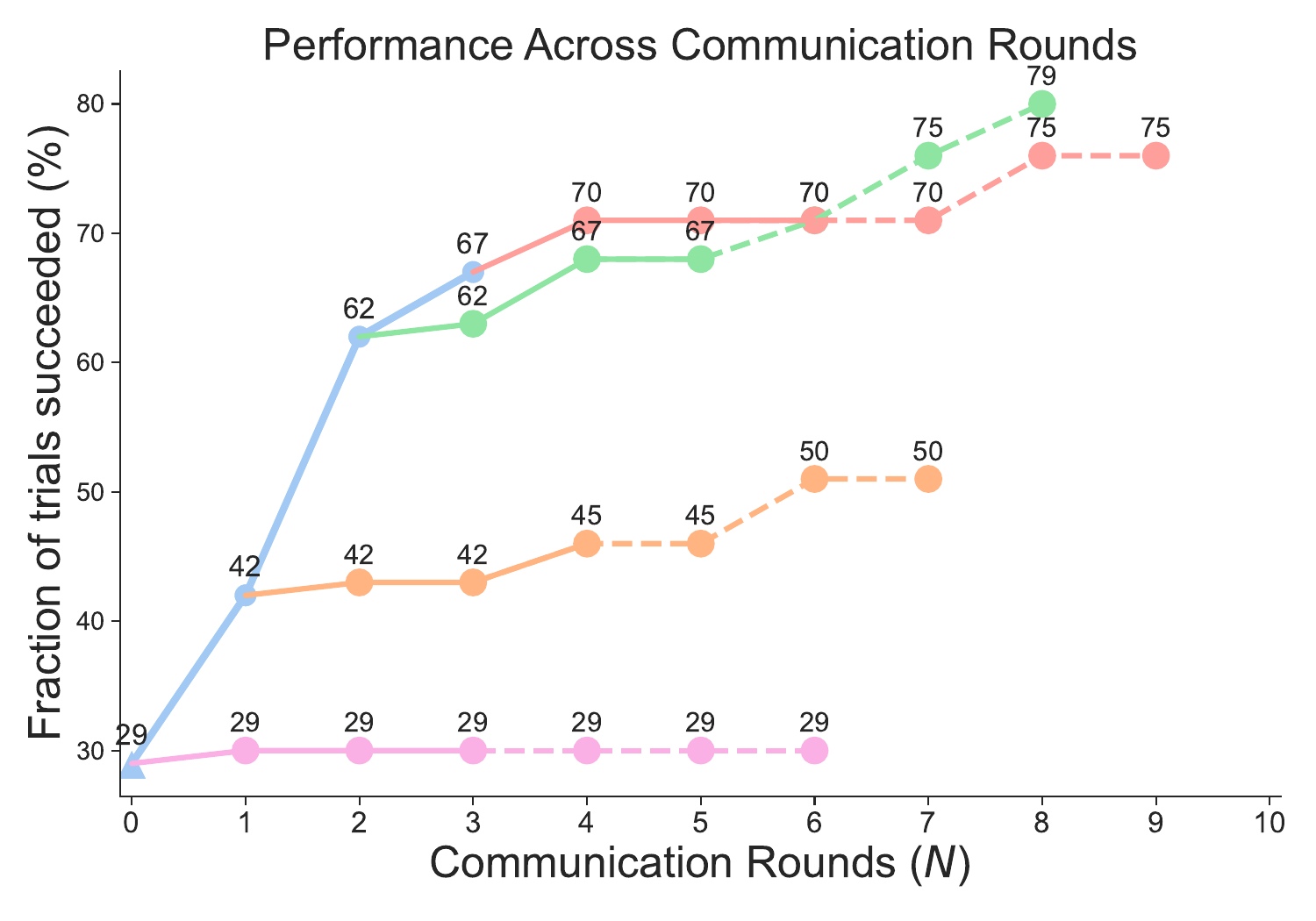}
    \captionof{figure}{\textbf{Collaborative Setting (Mixtral); Dirt Collection}. Blue: GPT-4–guided baseline. Dashed\quad(– –): extra collaboration rounds.}
    \label{fig:weak-weak}
\end{minipage}
\vspace{-5pt}
\end{figure*}

\paragraph{More Communication Rounds Improve Performance}Since communication is central to our framework, we examine the benefits of multiple rounds per task trial, assessing communication and perspective-taking amid escalating opportunities. We pair a GPT-4 \collabvoyager expert with a Mixtral-8x7B \collabvoyager weak agent. \Cref{fig:comm-rounds} demonstrates communication's value through a sharp rise in task completion rates with additional rounds, while \Cref{fig:beliefs-evolution-rounds} illustrates the expert's progressively accurate grasp of the partner's mental state. Further evidence in \Cref{appendix:comm-experiments} affirms communication's importance in instructive and collaborative settings, even for demanding Minecraft tasks (``Craft a pickaxe'' and ``Mine Iron'') involving \collabvoyager agents.

\paragraph{Using a less constrained communication protocol}

We introduce the \collabvoyager communication protocol in \Cref{ssec:comm-medium}, initiating agent conversation only after task failure. To demonstrate the framework's flexibility, we test two \collabvoyager tasks with a dynamic communication setup, allowing the weak agent to choose whether to initiate communication when uncertain or attempt the task independently. \Cref{tab:flexible-communication} (\Cref{appendix:additional}) reveals comparable task completion rates between the default and flexible protocols, underscoring \collabvoyager's independence from specific communication designs. Qualitatively, weaker models like Mistral-7B consistently act confidently, never seeking help despite the option, while both models adeptly bypass assistance for tasks aligned with prior experience.

\begin{table*}[t]
\scriptsize
\centering
\caption{We follow Voyager's methodology \citep{wang2023voyager} for reporting the Minecraft tech-tree results. The reported scores represent the number of prompting iterations until the milestone was accomplished, averaged over three trials. The fewer the iterations, the more efficient the method. $N/A$ represents the agent being unable to solve the task. \greentext{Best performance} in terms of iterations required and number of times milestone is reached (out of $3$ trials).}
\label{table:tech_tree}
{\tabcolsep=8pt\def\arraystretch{1.1}
\begin{tabularx}{\textwidth}{l l *{4}{X}}
\toprule
\textbf{Method} & \textbf{Model} & \textbf{Wooden Tool} & \textbf{Stone Tool} & \textbf{Iron Tool} \\
\midrule
Voyager~\citeyearpar{wang2023voyager}
& GPT-4 & $6 \pm$ \tiny{$2$} $(\sfrac{3}{3})$ & $11 \pm$\tiny{$2$} $(\sfrac{3}{3})$ & $21 \pm$\tiny{$7$} $(\sfrac{3}{3})$ \\
ReAct~\citeyearpar{wang2023voyager} & GPT-4 & N/A $(\sfrac{0}{3})$ & N/A $(\sfrac{0}{3})$ & N/A $(\sfrac{0}{3})$\\
Reflexion~\citeyearpar{wang2023voyager} & GPT-4 & N/A $(\sfrac{0}{3})$ & N/A $(\sfrac{0}{3})$&N/A $(\sfrac{0}{3})$\\
AutoGPT~\citeyearpar{wang2023voyager} & GPT-4&  $92 \pm$\tiny{$72$} ${(\sfrac{3}{3})}$ & $92 \pm$\tiny{$72$} ${(\sfrac{3}{3})}$ & $135 \pm$\tiny{$103$} ${(\sfrac{3}{3})}$\\

\midrule
\multirow{3}{*}{\makecell[l]{Voyager \\ (open-weight)}}
& Mixtral 8$\times$7B & N/A $(\sfrac{0}{3})$ & N/A $(\sfrac{0}{3})$ & N/A $(\sfrac{0}{3})$ \\
& Llama 3.1-8B & N/A $(\sfrac{0}{3})$ & N/A $(\sfrac{0}{3})$ & N/A $(\sfrac{0}{3})$ \\
& Llama 3.1-70B & $\bestscore{19\pm 9}$ $\bestscore{(\sfrac{3}{3})}$ & N/A $(\sfrac{0}{3})$ & N/A $(\sfrac{0}{3})$ \\
\midrule
\multirow{3}{*}{\makecell[l]{\collabvoyager \\ (w/o collaboration)}}
& Mixtral 8$\times$7B & $86 \pm$\tiny{$17$}$\bestscore{(\sfrac{3}{3})}$ & N/A $(\sfrac{0}{3})$ & N/A $(\sfrac{0}{3})$ \\
& Llama 3.1-8B & $120 \pm$\tiny{$0$} ${(\sfrac{1}{3})}$ & N/A $(\sfrac{0}{3})$ & N/A $(\sfrac{0}{3})$ \\
& Llama 3.1-70B & $\bestscore{50}$\tiny{$\bestscore{\pm17}$} $\bestscore{(\sfrac{3}{3})}$ & $\bestscore{113}$\tiny{$\bestscore{\pm24}$} $\bestscore{(\sfrac{2}{3})}$ & N/A $(\sfrac{0}{3})$ \\
\midrule
\multirow{3}{*}{\makecell[l]{\collabvoyager \\ (mixed-expertise collaboration)}}
& Mixtral 8$\times$7B & $\bestscore{51}$\tiny{$\bestscore{\pm 20}$} $\bestscore{(\sfrac{3}{3})}$ & N/A $(\sfrac{0}{3})$ & N/A $(\sfrac{0}{3})$ \\
& Llama 3.1-8B & ${100}$ \tiny{${\pm 30}$} $\bestscore{(\sfrac{3}{3})}$ & N/A $(\sfrac{0}{3})$ & N/A $(\sfrac{0}{3})$ \\
& Llama 3.1-70B & ${62}$\tiny{${\pm 14}$} $\bestscore{(\sfrac{3}{3})}$& $\bestscore{79}$\tiny{$\bestscore{\pm 25}$}$\bestscore{(\sfrac{3}{3})}$ & $\bestscore{120}$\tiny{$\bestscore{\pm 30}$} $\bestscore{(\sfrac{2}{3})}$\\
\midrule
\makecell[l]{\collabvoyager \\ (mixed-expertise collab w/ Llama 3.1-70B)}
& Mixtral 8$\times$7B & $\bestscore{65}$\tiny{$\bestscore{\pm 12}$} $(\sfrac{3}{3})$ & N/A $(\sfrac{0}{3})$ & N/A $(\sfrac{0}{3})$ \\
\bottomrule
\end{tabularx}}
\end{table*}

\subsection{\collabvoyager Lifelong Learning}

We evaluate \collabvoyager's lifelong learning capabilities against both GPT-4 and open-weight Voyager \citep{wang2023voyager} variants. The progression through increasingly complex technological achievements (from wooden to diamond tools) provides a rigorous test-bed for assessing lifelong learning and collaboration abilities of \collabvoyager. We follow the methodology from Voyager and evaluate performance on three key milestones: (1) \textit{wooden tools}, (2) \textit{stone tools} and (3) \textit{iron tools}.

In line with the findings in \Cref{table:voyager-base}, \Cref{table:tech_tree} highlights the performance gap between GPT-4 powered Voyager and its open-weight variants, with only Llama 3.1-70B \citep{dubey2024llama} reaching the \textit{wooden tools} milestone. On the other hand, \collabvoyager shows substantial improvement across the board both in terms of achieved milestones and efficiency (number of iterations). Notably, Llama 3.1-70B \citep{dubey2024llama} powered \collabvoyager achieves two additional tech-tree milestones, nearing the performance with the original GPT-4 powered Voyager \citep{wang2023voyager}. We observe a steady increase in efficiency (number of iterations) as the size of the LLM increases: Llama 3.1-70B \citep{dubey2024llama} powered \collabvoyager is $1.62\times$ more efficient than its Llama 3.1-8B counterpart for reaching the \textit{wooden tools} milestone, suggesting  that larger models can better utilize the collaborative learning framework. Furthermore, to understand the impact of collaboration in a mixed-expertise collaborative setting, we experiment with a non-collaborative \collabvoyager ablation. Notably, \collabvoyager agents without collaboration require on average $15$ more prompting iterations to reach the \textit{wooden tools} milestone and $34$ iterations to reach the \textit{stone tools} milestone. Thus, the collaborative framework not only enhances task completion rates but also enables more efficient acquisition of complex skills. We further test the generalization capabilities of \collabvoyager agents in \Cref{appendix:post-collab} where we show how even after collaboration finishes, the agents can reuse the knowledge in out-of-distribution tasks.

\subsection{\collabvoyager Collaborative Learning Setting}
\label{sec:colab-learning}

While prior sections focused on instructive learning from stronger peers, we now consider a more challenging setup: collaboration between two \collabvoyager agents of \textbf{matched capabilities}, each powered by the same type of open-weight LLM. This setting reflects mutual cultural learning without access to an expert where the agents must bootstrap improvements solely through interaction, thereby mimicking certain qualities of emergence.
Critically, we evaluate not individual trajectories, but instead focus on \textbf{population-level performance}:  after \(N\) rounds of communication, what fraction of agent pairs from the total pool succeed in completing the task? As shown in~\Cref{table:main}, when agents begin from unprimed states, collaboration often leads to stagnation or even degradation which mirrors the \textit{``blind leading the blind''} dynamic, where flawed mental models are mutually reinforced. 

However, even minimal prior expert instruction—such as a single round of GPT-4–guided collaboration—can prime agents for more effective peer interaction, even if they still fail when acting alone. As shown in~\Cref{fig:weak-weak}, we pair Mixtral-8x7B agents that each received \(N\) communication rounds of GPT-4 guided trial before interacting. We observe  that the fraction of successful agents steadily increases with additional communication rounds between themselves (Mixtral agents). From a \(62\%\) baseline population succeeding, we reach \(79\%\)  of the population succeeding after seven turns, showing a \(17\%\) increment. This result reflects a form of test-time compute scaling, where richer interaction, not larger models, drives the emergent improvements across the group.

This also aligns with the Condorcet Jury Theorem~\citep{austen1996information}, which predicts that collective decisions improve when individuals exceed a baseline level of competence. Here, we see a distributed variant of this principle in action: once agents cross a minimal competence threshold, the communication induced bootstrapping can yield reliable gains at the population level.



\section{Conclusion}\label{sec:conclusion}

The \collabvoyager framework demonstrates that cultural learning can overcome the limitations of open-weight LLMs in embodied environments. Integrating theory of mind, natural communication, and structured memory systems enables \collabvoyager agents to learn from both mixed and matched expertise peers and overcome fundamental failure modes in perception and code generation. Our experiments show that collaboration significantly improves performance on lifelong learning scenarios, with agents collecting $2.3\times$ more unique items and achieving $3\times$ more tech-tree milestones when paired with expert agents. Crucially, in the collaborative setting, we show that the fraction of successful agents scales as a function of communication rounds; revealing emergent gains through perspective taking and interaction. These results suggest that social learning could reduce embodied AI's dependence on large proprietary models, opening new paths for democratized agent development aligned with Green AI principles \citep{schwartzGreenAI2019}.

\begin{ack}
We thank OpenAI's Researcher Access Program for providing API credits that enabled more comprehensive experiments. We are also grateful to Dr. Max van Duijn for his constructive critique and insights on Theory of Mind, to Dr. Jesse Krijthe for his thoughtful input on the Condorcet Jury Theorem, and to members of the Pattern Recognition Lab at TU Delft for their valuable feedback. 
\end{ack}

\bibliographystyle{unsrtnat} 
\bibliography{references}    

\newpage
\section*{NeurIPS Paper Checklist}

\begin{enumerate}

\item {\bf Claims}
    \item[] Question: Do the main claims made in the abstract and introduction accurately reflect the paper's contributions and scope?
    \item[] Answer: \answerYes{} 
    \item[] Justification: We back up the claims made in the abstract and introduction in \Cref{sec:experiments}, where we run extensive experiments in Minecraft and show the benefits of the proposed framework. Furthermore, we offer qualitative results through various figures and text snippets in the manuscript. These are available in the main paper as well as in the appendix (e.g. \Cref{appendix:sample-conv}).
    \item[] Guidelines:
    \begin{itemize}
        \item The answer NA means that the abstract and introduction do not include the claims made in the paper.
        \item The abstract and/or introduction should clearly state the claims made, including the contributions made in the paper and important assumptions and limitations. A No or NA answer to this question will not be perceived well by the reviewers. 
        \item The claims made should match theoretical and experimental results, and reflect how much the results can be expected to generalize to other settings. 
        \item It is fine to include aspirational goals as motivation as long as it is clear that these goals are not attained by the paper. 
    \end{itemize}

\item {\bf Limitations}
    \item[] Question: Does the paper discuss the limitations of the work performed by the authors?
    \item[] Answer: \answerNo{} 
    \item[] Justification: 
    We acknowledge the limitations of \collabvoyager and we discuss them throughout the paper when necessary. That said we do not have a standalone Limitations section.
    \item[] Guidelines:
    \begin{itemize}
        \item The answer NA means that the paper has no limitation while the answer No means that the paper has limitations, but those are not discussed in the paper. 
        \item The authors are encouraged to create a separate "Limitations" section in their paper.
        \item The paper should point out any strong assumptions and how robust the results are to violations of these assumptions (e.g., independence assumptions, noiseless settings, model well-specification, asymptotic approximations only holding locally). The authors should reflect on how these assumptions might be violated in practice and what the implications would be.
        \item The authors should reflect on the scope of the claims made, e.g., if the approach was only tested on a few datasets or with a few runs. In general, empirical results often depend on implicit assumptions, which should be articulated.
        \item The authors should reflect on the factors that influence the performance of the approach. For example, a facial recognition algorithm may perform poorly when image resolution is low or images are taken in low lighting. Or a speech-to-text system might not be used reliably to provide closed captions for online lectures because it fails to handle technical jargon.
        \item The authors should discuss the computational efficiency of the proposed algorithms and how they scale with dataset size.
        \item If applicable, the authors should discuss possible limitations of their approach to address problems of privacy and fairness.
        \item While the authors might fear that complete honesty about limitations might be used by reviewers as grounds for rejection, a worse outcome might be that reviewers discover limitations that aren't acknowledged in the paper. The authors should use their best judgment and recognize that individual actions in favor of transparency play an important role in developing norms that preserve the integrity of the community. Reviewers will be specifically instructed to not penalize honesty concerning limitations.
    \end{itemize}

\item {\bf Theory assumptions and proofs}
    \item[] Question: For each theoretical result, does the paper provide the full set of assumptions and a complete (and correct) proof?
    \item[] Answer: \answerNA{} 
    \item[] Justification: The manuscript does not include any theoretical result.
    \item[] Guidelines:
    \begin{itemize}
        \item The answer NA means that the paper does not include theoretical results. 
        \item All the theorems, formulas, and proofs in the paper should be numbered and cross-referenced.
        \item All assumptions should be clearly stated or referenced in the statement of any theorems.
        \item The proofs can either appear in the main paper or the supplemental material, but if they appear in the supplemental material, the authors are encouraged to provide a short proof sketch to provide intuition. 
        \item Inversely, any informal proof provided in the core of the paper should be complemented by formal proofs provided in appendix or supplemental material.
        \item Theorems and Lemmas that the proof relies upon should be properly referenced. 
    \end{itemize}

    \item {\bf Experimental result reproducibility}
    \item[] Question: Does the paper fully disclose all the information needed to reproduce the main experimental results of the paper to the extent that it affects the main claims and/or conclusions of the paper (regardless of whether the code and data are provided or not)?
    \item[] Answer: \answerYes{}
    \item[] Justification: Since our method relies on LLMs, we provide a full comprehensive list of the models used (via API), the underlying prompts used to query the LLM as well as extensive details in \Cref{sec:experiments} that allows for potential reproducibility. Importantly, our method uses significant scaffolding to enable multi-agent communication in Minecraft. While we do not include code snippets in the manuscript, we provide the full code as part of the Supplementary Materials.
    \item[] Guidelines:
    \begin{itemize}
        \item The answer NA means that the paper does not include experiments.
        \item If the paper includes experiments, a No answer to this question will not be perceived well by the reviewers: Making the paper reproducible is important, regardless of whether the code and data are provided or not.
        \item If the contribution is a dataset and/or model, the authors should describe the steps taken to make their results reproducible or verifiable. 
        \item Depending on the contribution, reproducibility can be accomplished in various ways. For example, if the contribution is a novel architecture, describing the architecture fully might suffice, or if the contribution is a specific model and empirical evaluation, it may be necessary to either make it possible for others to replicate the model with the same dataset, or provide access to the model. In general. releasing code and data is often one good way to accomplish this, but reproducibility can also be provided via detailed instructions for how to replicate the results, access to a hosted model (e.g., in the case of a large language model), releasing of a model checkpoint, or other means that are appropriate to the research performed.
        \item While NeurIPS does not require releasing code, the conference does require all submissions to provide some reasonable avenue for reproducibility, which may depend on the nature of the contribution. For example
        \begin{enumerate}
            \item If the contribution is primarily a new algorithm, the paper should make it clear how to reproduce that algorithm.
            \item If the contribution is primarily a new model architecture, the paper should describe the architecture clearly and fully.
            \item If the contribution is a new model (e.g., a large language model), then there should either be a way to access this model for reproducing the results or a way to reproduce the model (e.g., with an open-source dataset or instructions for how to construct the dataset).
            \item We recognize that reproducibility may be tricky in some cases, in which case authors are welcome to describe the particular way they provide for reproducibility. In the case of closed-source models, it may be that access to the model is limited in some way (e.g., to registered users), but it should be possible for other researchers to have some path to reproducing or verifying the results.
        \end{enumerate}
    \end{itemize}

\item {\bf Open access to data and code}
    \item[] Question: Does the paper provide open access to the data and code, with sufficient instructions to faithfully reproduce the main experimental results, as described in supplemental material?
    \item[] Answer: \answerYes{} 
    \item[] Justification: We provide the full code repository used to execute any experiment referenced in the manuscript in the Supplementary Material. We anonymize the code and provide adequate instructions inside the README for reproducibility.
    \item[] Guidelines:
    \begin{itemize}
        \item The answer NA means that paper does not include experiments requiring code.
        \item Please see the NeurIPS code and data submission guidelines (\url{https://nips.cc/public/guides/CodeSubmissionPolicy}) for more details.
        \item While we encourage the release of code and data, we understand that this might not be possible, so “No” is an acceptable answer. Papers cannot be rejected simply for not including code, unless this is central to the contribution (e.g., for a new open-source benchmark).
        \item The instructions should contain the exact command and environment needed to run to reproduce the results. See the NeurIPS code and data submission guidelines (\url{https://nips.cc/public/guides/CodeSubmissionPolicy}) for more details.
        \item The authors should provide instructions on data access and preparation, including how to access the raw data, preprocessed data, intermediate data, and generated data, etc.
        \item The authors should provide scripts to reproduce all experimental results for the new proposed method and baselines. If only a subset of experiments are reproducible, they should state which ones are omitted from the script and why.
        \item At submission time, to preserve anonymity, the authors should release anonymized versions (if applicable).
        \item Providing as much information as possible in supplemental material (appended to the paper) is recommended, but including URLs to data and code is permitted.
    \end{itemize}

\item {\bf Experimental setting/details}
    \item[] Question: Does the paper specify all the training and test details (e.g., data splits, hyperparameters, how they were chosen, type of optimizer, etc.) necessary to understand the results?
    \item[] Answer: \answerYes{} 
    \item[] Justification: Since the method to interact with LLMs is through text prompts, we provide these artifacts together with how we set up the experiments. These details can be found in \Cref{sec:experiments}. The full details (code) are provided in the Supplementary Material.
    \item[] Guidelines:
    \begin{itemize}
        \item The answer NA means that the paper does not include experiments.
        \item The experimental setting should be presented in the core of the paper to a level of detail that is necessary to appreciate the results and make sense of them.
        \item The full details can be provided either with the code, in appendix, or as supplemental material.
    \end{itemize}

\item {\bf Experiment statistical significance}
    \item[] Question: Does the paper report error bars suitably and correctly defined or other appropriate information about the statistical significance of the experiments?
    \item[] Answer: \answerYes{}
    \item[] Justification: We report error bars where possible throughout this manuscript. Specifically, we provide error bars for the Minecraft tech-tree experiments in \Cref{table:tech_tree}. Moreover, to ensure proper statistical significance, we perform a large number of individual trials across all the experiments in \Cref{sec:experiments}.
    \item[] Guidelines:
    \begin{itemize}
        \item The answer NA means that the paper does not include experiments.
        \item The authors should answer "Yes" if the results are accompanied by error bars, confidence intervals, or statistical significance tests, at least for the experiments that support the main claims of the paper.
        \item The factors of variability that the error bars are capturing should be clearly stated (for example, train/test split, initialization, random drawing of some parameter, or overall run with given experimental conditions).
        \item The method for calculating the error bars should be explained (closed form formula, call to a library function, bootstrap, etc.)
        \item The assumptions made should be given (e.g., Normally distributed errors).
        \item It should be clear whether the error bar is the standard deviation or the standard error of the mean.
        \item It is OK to report 1-sigma error bars, but one should state it. The authors should preferably report a 2-sigma error bar than state that they have a 96\% CI, if the hypothesis of Normality of errors is not verified.
        \item For asymmetric distributions, the authors should be careful not to show in tables or figures symmetric error bars that would yield results that are out of range (e.g. negative error rates).
        \item If error bars are reported in tables or plots, The authors should explain in the text how they were calculated and reference the corresponding figures or tables in the text.
    \end{itemize}

\item {\bf Experiments compute resources}
    \item[] Question: For each experiment, does the paper provide sufficient information on the computer resources (type of compute workers, memory, time of execution) needed to reproduce the experiments?
    \item[] Answer: \answerYes{} 
    \item[] Justification: While our paper is not based on direct compute resources like GPUs, we do acknowledge in \Cref{appendix:llm} the cloud providers used throughout the experiments.
    \item[] Guidelines:
    \begin{itemize}
        \item The answer NA means that the paper does not include experiments.
        \item The paper should indicate the type of compute workers CPU or GPU, internal cluster, or cloud provider, including relevant memory and storage.
        \item The paper should provide the amount of compute required for each of the individual experimental runs as well as estimate the total compute. 
        \item The paper should disclose whether the full research project required more compute than the experiments reported in the paper (e.g., preliminary or failed experiments that didn't make it into the paper). 
    \end{itemize}
    
\item {\bf Code of ethics}
    \item[] Question: Does the research conducted in the paper conform, in every respect, with the NeurIPS Code of Ethics \url{https://neurips.cc/public/EthicsGuidelines}?
    \item[] Answer: \answerYes{} 
    \item[] Justification: We have followed the NeurIPS Code of Ethics while writing and working at this manuscript.
    \item[] Guidelines:
    \begin{itemize}
        \item The answer NA means that the authors have not reviewed the NeurIPS Code of Ethics.
        \item If the authors answer No, they should explain the special circumstances that require a deviation from the Code of Ethics.
        \item The authors should make sure to preserve anonymity (e.g., if there is a special consideration due to laws or regulations in their jurisdiction).
    \end{itemize}

\item {\bf Broader impacts}
    \item[] Question: Does the paper discuss both potential positive societal impacts and negative societal impacts of the work performed?
    \item[] Answer: \answerNo{}
    \item[] Justification:  While the \collabvoyager framework is designed to work with both humans and agents, we believe there is no foreseeable negative societal impact that can stem from this specific manuscript. However, we do mention that our framework might decrease the dependence on proprietary models and thus affect in a positive way global problems like climate change.
    \item[] Guidelines:
    \begin{itemize}
        \item The answer NA means that there is no societal impact of the work performed.
        \item If the authors answer NA or No, they should explain why their work has no societal impact or why the paper does not address societal impact.
        \item Examples of negative societal impacts include potential malicious or unintended uses (e.g., disinformation, generating fake profiles, surveillance), fairness considerations (e.g., deployment of technologies that could make decisions that unfairly impact specific groups), privacy considerations, and security considerations.
        \item The conference expects that many papers will be foundational research and not tied to particular applications, let alone deployments. However, if there is a direct path to any negative applications, the authors should point it out. For example, it is legitimate to point out that an improvement in the quality of generative models could be used to generate deepfakes for disinformation. On the other hand, it is not needed to point out that a generic algorithm for optimizing neural networks could enable people to train models that generate Deepfakes faster.
        \item The authors should consider possible harms that could arise when the technology is being used as intended and functioning correctly, harms that could arise when the technology is being used as intended but gives incorrect results, and harms following from (intentional or unintentional) misuse of the technology.
        \item If there are negative societal impacts, the authors could also discuss possible mitigation strategies (e.g., gated release of models, providing defenses in addition to attacks, mechanisms for monitoring misuse, mechanisms to monitor how a system learns from feedback over time, improving the efficiency and accessibility of ML).
    \end{itemize}
    
\item {\bf Safeguards}
    \item[] Question: Does the paper describe safeguards that have been put in place for responsible release of data or models that have a high risk for misuse (e.g., pretrained language models, image generators, or scraped datasets)?
    \item[] Answer: \answerNA{} 
    \item[] Justification: Our paper does not pose such risks.
    \item[] Guidelines:
    \begin{itemize}
        \item The answer NA means that the paper poses no such risks.
        \item Released models that have a high risk for misuse or dual-use should be released with necessary safeguards to allow for controlled use of the model, for example by requiring that users adhere to usage guidelines or restrictions to access the model or implementing safety filters. 
        \item Datasets that have been scraped from the Internet could pose safety risks. The authors should describe how they avoided releasing unsafe images.
        \item We recognize that providing effective safeguards is challenging, and many papers do not require this, but we encourage authors to take this into account and make a best faith effort.
    \end{itemize}

\item {\bf Licenses for existing assets}
    \item[] Question: Are the creators or original owners of assets (e.g., code, data, models), used in the paper, properly credited and are the license and terms of use explicitly mentioned and properly respected?
    \item[] Answer: \answerYes{} 
    \item[] Justification: We credit the LLM providers in \Cref{appendix:llm} both for crediting their work and helping the reader reproduce the results in the paper.
    \item[] Guidelines:
    \begin{itemize}
        \item The answer NA means that the paper does not use existing assets.
        \item The authors should cite the original paper that produced the code package or dataset.
        \item The authors should state which version of the asset is used and, if possible, include a URL.
        \item The name of the license (e.g., CC-BY 4.0) should be included for each asset.
        \item For scraped data from a particular source (e.g., website), the copyright and terms of service of that source should be provided.
        \item If assets are released, the license, copyright information, and terms of use in the package should be provided. For popular datasets, \url{paperswithcode.com/datasets} has curated licenses for some datasets. Their licensing guide can help determine the license of a dataset.
        \item For existing datasets that are re-packaged, both the original license and the license of the derived asset (if it has changed) should be provided.
        \item If this information is not available online, the authors are encouraged to reach out to the asset's creators.
    \end{itemize}

\item {\bf New assets}
    \item[] Question: Are new assets introduced in the paper well documented and is the documentation provided alongside the assets?
    \item[] Answer: \answerNA{} 
    \item[] Justification: Since \collabvoyager makes use of existing open-weight LLMs, our paper does not release any new assets.
    \item[] Guidelines:
    \begin{itemize}
        \item The answer NA means that the paper does not release new assets.
        \item Researchers should communicate the details of the dataset/code/model as part of their submissions via structured templates. This includes details about training, license, limitations, etc. 
        \item The paper should discuss whether and how consent was obtained from people whose asset is used.
        \item At submission time, remember to anonymize your assets (if applicable). You can either create an anonymized URL or include an anonymized zip file.
    \end{itemize}

\item {\bf Crowdsourcing and research with human subjects}
    \item[] Question: For crowdsourcing experiments and research with human subjects, does the paper include the full text of instructions given to participants and screenshots, if applicable, as well as details about compensation (if any)? 
    \item[] Answer: \answerNA{} 
    \item[] Justification: The manuscript and research process does not involve crowdsourcing nor research with human subjects. While \Cref{table:main} showcases a feasibility study with a human expert, the human is one of the authors.
    \item[] Guidelines:
    \begin{itemize}
        \item The answer NA means that the paper does not involve crowdsourcing nor research with human subjects.
        \item Including this information in the supplemental material is fine, but if the main contribution of the paper involves human subjects, then as much detail as possible should be included in the main paper. 
        \item According to the NeurIPS Code of Ethics, workers involved in data collection, curation, or other labor should be paid at least the minimum wage in the country of the data collector. 
    \end{itemize}

\item {\bf Institutional review board (IRB) approvals or equivalent for research with human subjects}
    \item[] Question: Does the paper describe potential risks incurred by study participants, whether such risks were disclosed to the subjects, and whether Institutional Review Board (IRB) approvals (or an equivalent approval/review based on the requirements of your country or institution) were obtained?
    \item[] Answer: \answerNA{} 
    \item[] Justification: The manuscript and research process does not involve crowdsourcing nor research with human subjects. While \Cref{table:main} showcases a feasibility study with a human expert, the human is one of the authors.
    \item[] Guidelines:
    \begin{itemize}
        \item The answer NA means that the paper does not involve crowdsourcing nor research with human subjects.
        \item Depending on the country in which research is conducted, IRB approval (or equivalent) may be required for any human subjects research. If you obtained IRB approval, you should clearly state this in the paper. 
        \item We recognize that the procedures for this may vary significantly between institutions and locations, and we expect authors to adhere to the NeurIPS Code of Ethics and the guidelines for their institution. 
        \item For initial submissions, do not include any information that would break anonymity (if applicable), such as the institution conducting the review.
    \end{itemize}

\item {\bf Declaration of LLM usage}
    \item[] Question: Does the paper describe the usage of LLMs if it is an important, original, or non-standard component of the core methods in this research? Note that if the LLM is used only for writing, editing, or formatting purposes and does not impact the core methodology, scientific rigorousness, or originality of the research, declaration is not required.
    \item[] Answer: \answerYes{} 
    \item[] Justification: We do not use LLMs to formulate our research problem however LLMs are being used for the agents that we are proposing. 
    \item[] Guidelines:
    \begin{itemize}
        \item The answer NA means that the core method development in this research does not involve LLMs as any important, original, or non-standard components.
        \item Please refer to our LLM policy (\url{https://neurips.cc/Conferences/2025/LLM}) for what should or should not be described.
    \end{itemize}

\end{enumerate}

\newpage
\input{appendix}


\end{document}

%% file: appendix.tex
\appendix

\begin{center}
\Large
\textbf{MindForge: Empowering Embodied Agents with Theory of Mind for Lifelong
Cultural Learning} \\
\vspace{0.2em}Appendices \\
\vspace{0.3em}
\end{center}

\section{Additional Experimental Results}
\label{appendix:additional-experiments}

\subsection{Collaborative Experiments}
\label{appendix:comm-experiments}

We provide an extended set of experiments to reinforce our findings in \Cref{fig:comm-rounds} and \Cref{fig:weak-weak}. Specifically, we consider atomic tasks that are significantly harder than dirt and wood collection in order to demonstrate the generality of our framework: (1) mining an iron block and (2) crafting a wooden pickaxe.

\begin{table}[htb]
    \scriptsize
  \caption{Agent performance increases as the number of rounds grows in both collaborative and instructive settings. In the instructive setting, we use Llama 3.1-70B as the weak agent and GPT-4 as the strong agent. Similarly, collaborative experiments use Llama 3.1-70B for both \collabvoyager agents.}
    {\tabcolsep=9pt\def\arraystretch{1.1}
    \begin{tabularx}{\columnwidth}{@{}>{\arraybackslash}l *5{X}}
      \toprule
      \textbf{Setting} & \textbf{Task} & \textbf{Comm. Round 0} &\textbf{Comm. Round 1} & \textbf{Comm. Round 2} & \textbf{Comm. Round 3} \\ 
      \midrule
      Instructive & Mine Iron & $41\%$ & $50\%$ & $54\%$ & $62\%$ \\
      Instructive & Craft a Pickaxe & $20\%$ & $33\%$ & $41\%$ &$45\%$ \\
      \midrule
      Collaborative & Craft a Pickaxe & $20\%$ & $25\%$ & $33\%$ & $33\%$ \\
      \bottomrule
    \end{tabularx}}
    \label{table:comm-experiments-appendix}
\end{table}

As \Cref{table:comm-experiments-appendix} shows, extend communication and perspective-taking between \collabvoyager agents enable significant improvements on complex Minecraft tasks that require compositionality ($41\% \rightarrow 62\%$ and $20\% \rightarrow 45\%$). Similarly to \Cref{fig:weak-weak}, we see that a purely collaborative setting results in smaller performance gains ($20\% \rightarrow 33\%$) than instructive learning, due to reinforcing wrong beliefs and similar thinking patterns.

\subsection{Post-Collaboration and OOD tasks}
\label{appendix:post-collab}

Do social interactions have long-term effects on agents and help them perform better on different tasks? We consider the impact of interactions in post-collaboration and out-of-distribution scenarios to assess \collabvoyager's generalization and lifelong learning capabilities. Specifically, we define post-collaboration scenarios as a non-collaborative setting where the \collabvoyager agent tries to complete a task based on prior beliefs formed from a previous collaborative experience stored in the semantic memory. Moreover, we consider scenarios where the agent is tasked with solving a novel task with prior collaborative experience on an adjacent task in the Minecraft tech-tree. As \Cref{table:ood} shows, collaboration between \collabvoyager agents together with semantic memory showcase an average improvement of $8.33\%$ for in-distribution tasks (dirt collection) together with an average improvement of $8.78\%$ in non-collaborative out-of-distribution scenarios (wood collection). These results showcase the importance of semantic memory in achieving continual learning through collaboration. 

\begin{table}[h]
\caption{Post-collaboration and out-of-distribution generalization of the \collabvoyager agent. We report the task completion rates across $24$ individual trials.}
  \label{table:ood}
  \scriptsize
  {\tabcolsep=10pt\def\arraystretch{1.1}
  \begin{tabularx}{\columnwidth}{@{}>{\arraybackslash}l *4{X}}
  \toprule
  \textbf{Scenario} & \textbf{Task} & \textbf{Mistral-7B} & \textbf{Mixtral-8x7B} & \textbf{Llama 3.1-8B}\\ 
  \midrule
  No Collaboration & Dirt & $37.5\%$ & $29.15\%$ & $12.5\%$ \\
  Post-Collaboration & Dirt & \bm{$37.5\%$} & \bm{$41.66\%$} & \bm{$25\%$}\\
  \midrule
  No Collaboration & Wood & $33.3\%$ & $50\%$ & $7\%$ \\
  OOD post-collaboration & Wood & \bm{$41.66\%$} & \bm{$58.33\%$}& \bm{$16.66\%$}\\
  \bottomrule
  \end{tabularx}}
\end{table}

\subsection{Alternative Communication Protocols}
\label{appendix:additional}
\Cref{tab:flexible-communication} shows the performance of \collabvoyager agents when they are allowed to begin a conversation at point compared to when they can only initiate a conversation upon failure.

\begin{table}[hbt]
    \centering
    \caption{Comparison between the default \collabvoyager communication protocol and a more flexible alternative where agents can initiate the conversation at any point. Both protocols achieve similar results across different model sizes.}
    \label{tab:flexible-communication}
    \scriptsize
    \def\arraystretch{1.2}
    \begin{tabular}{@{}lllcccc@{}}
    \toprule
    \multirow{2}{*}{Model} & \multirow{2}{*}{Task} & \multirow{2}{*}{Setting} & \multicolumn{4}{c}{Task Completion Rate} \\
    \cmidrule(l){4-7}
    & & & Round 0 & Round 1 & Round 2 & Round 3 \\
    \midrule
    \multirow{4}{*}{Mixtral-8x7B} & \multirow{2}{*}{Mine dirt} & MindForge w/ flexible communication & 37\% & 45\% & 67\% & 67\% \\
    & & MindForge & 29\% & 42\% & 62\% & 67\% \\
    \cmidrule(l){2-7}
    & \multirow{2}{*}{Mine dirt and wood} & MindForge w/ flexible communication & 75\% & 79\% & 79\% & 83\% \\
    & & MindForge & 75\% & 79\% & 79\% & 83\% \\
    \midrule
    \multirow{4}{*}{Mistral-7B} & \multirow{2}{*}{Mine dirt} & MindForge w/ flexible communication & 37\% & 42\% & 45\% & 54\% \\
    & & MindForge & 37\% & 42\% & 45\% & 54\% \\
    \cmidrule(l){2-7}
    & \multirow{2}{*}{Mine dirt and wood} & MindForge w/ flexible communication & 41\% & 45\% & 50\% & 50\% \\
    & & MindForge & 41\% & 45\% & 50\% & 50\% \\
    \bottomrule
    \end{tabular}
\end{table}

\section{Additional Ablations}\label{appx:additional-ablations}
\subsection{Perspective Taking} \label{appendix:perspective}

To quantitatively assess the effect of perspective-taking, we perform an ablation study where we attempt to solve a Minecraft task without perspective-taking during communication in an instructive learning setting. \Cref{tab:perf-perspective} showcases how the absence of perspective-taking leads to worse performance, irrespective of how much agents communicate. Moreover, as the agents use more communication rounds when trying to solve a task, taking perspective of the other's agent situation increases task-completion rate.

\begin{table}[htbp]
    \centering
    \caption{Perspective-taking ablation. We consider the Minecraft task of collecting a dirt block. Improvement is quantified as the fraction of agents that solve the task across 24 individual trials.}
    \label{tab:perf-perspective}
    \scriptsize
    \def\arraystretch{1.2}
    \begin{tabular}{@{}lcccc@{}}
    \toprule
    \multirow{2}{*}{Model Variant} & \multicolumn{4}{c}{Task Completion Rate} \\
    \cmidrule(l){2-5}
    & Round 0 & Round 1 & Round 2 & Round 3 \\
    \midrule
    \collabvoyager w/ \textbf{perspective-taking} & 29\% & 42\% & 61\% & 67\% \\
    \collabvoyager w/o \textbf{perspective-taking} & 29\% & 37\% & 50\% & 54\% \\
    \midrule
    Improvement & 0\% & +5\% & +11\% & +13\% \\
    \bottomrule
    \end{tabular}
\end{table}

We attribute the positive correlation between performance and usage of perspective-taking during communication to the ability of the \collabvoyager agent to provide more relevant and directed advice. This observation is also supported by the substantial increase in performance in the first two rounds as presented in \Cref{tab:perf-perspective}, where the teacher agent provides more insightful information as it gets to know the other's agent situation better. 

Additionally, we further ablate our structured representation of Theory of Mind by considering an unstructured perspective-taking 2-step prompt presented in Think Twice \cite{wilf2023think}. \Cref{tab:perf-perspective-unstructured} showcases that a structured ToM representation provides a meaningful advantage on complex Minecraft tasks.

\begin{table}[htbp]
    \centering
    \caption{Structured perspective-taking ablation with unstructured alternative. The unstructured ToM approach is inspired from Think Twice \cite{wilf2023think}. Using a structured belief-system results in higher performance on complex Minecraft tasks.}
    \label{tab:perf-perspective-unstructured}
    \scriptsize
    \def\arraystretch{1.2}
    \begin{tabular}{@{}llcccc@{}}
    \toprule
    \multirow{2}{*}{Task} & \multirow{2}{*}{Model Variant} & \multicolumn{4}{c}{Task Completion Rate} \\
    \cmidrule(l){3-6}
    & & Round 0 & Round 1 & Round 2 & Round 3 \\
    \midrule
    Craft a Pickaxe & \collabvoyager & 20\% & \textbf{33}\% & \textbf{41}\% & \textbf{45}\% \\
    Craft a Pickaxe & \collabvoyager w/o structured ToM & 20\% & 29\% & 33\% & 41\% \\
    \midrule
    Mine Iron & \collabvoyager & 41\% & 50\% & 54\% & \textbf{62}\% \\
    Mine Iron & \collabvoyager w/o structured ToM & 41\% & 50\% & 54\% & 58\% \\
    \bottomrule
    \end{tabular}
\end{table}

\subsection{Memory Components} \label{appendix:memory}

\collabvoyager employs three types of memory subsystems, as discussed in \Cref{ssec:memory-subsys}. These components ensure that the agent can reuse and recall relevant information about the task it is trying to complete from past experiences and interactions with other agents. The importance of procedural memory with respect to task completion is already studied in \citep{wang2023voyager}, concluding that reusing previous code snippets becomes increasingly important with the difficulty of the tasks. To understand the effect of episodic memory in the \collabvoyager framework, we perform an ablation on the episodic memory component within two Minecraft tasks. \Cref{table:episodic-ablation} showcases the drop in completion rate without the use of the episodic memory component. While we do not consider the tech-tree experiments here, we expect that considering past failure episodes leads to increasingly better results.

\begin{table}[H]
\caption{Ablation on the episodic memory component in \collabvoyager on a single task. We report the task completion rates across 24 individual trials.}
\label{table:episodic-ablation}
\scriptsize
\def\arraystretch{1.2}
\begin{tabularx}{\columnwidth}{@{}>{\arraybackslash}l *3{Y}}
\toprule
\textbf{Scenario} & \textbf{Task} & \textbf{Mixtral-8x7B} & \textbf{Llama 3.1-8B}\\
\midrule
\collabvoyager w/ \textbf{episodic memory} & Dirt & $\bm{29.15\%}$ & $12.5\%$ \\ 
\collabvoyager w/o \textbf{episodic memory} & Dirt & $25\%$ & \bm{$16.66\%$}\\ 
\midrule
\collabvoyager w/ \textbf{episodic memory} & Wood & $\bm{50\%}$ & $\bm{7\%}$ \\ 
\collabvoyager w/o \textbf{episodic memory} & Wood & $45.83\%$& \bm{$7\%$}\\
\bottomrule
\end{tabularx}
\end{table}

\section{Implementation details}\label{appx:impl-details}

\subsection{Perspective Taking}
\begin{figure}[H]
\centering
\includegraphics[width=0.8\columnwidth, clip]{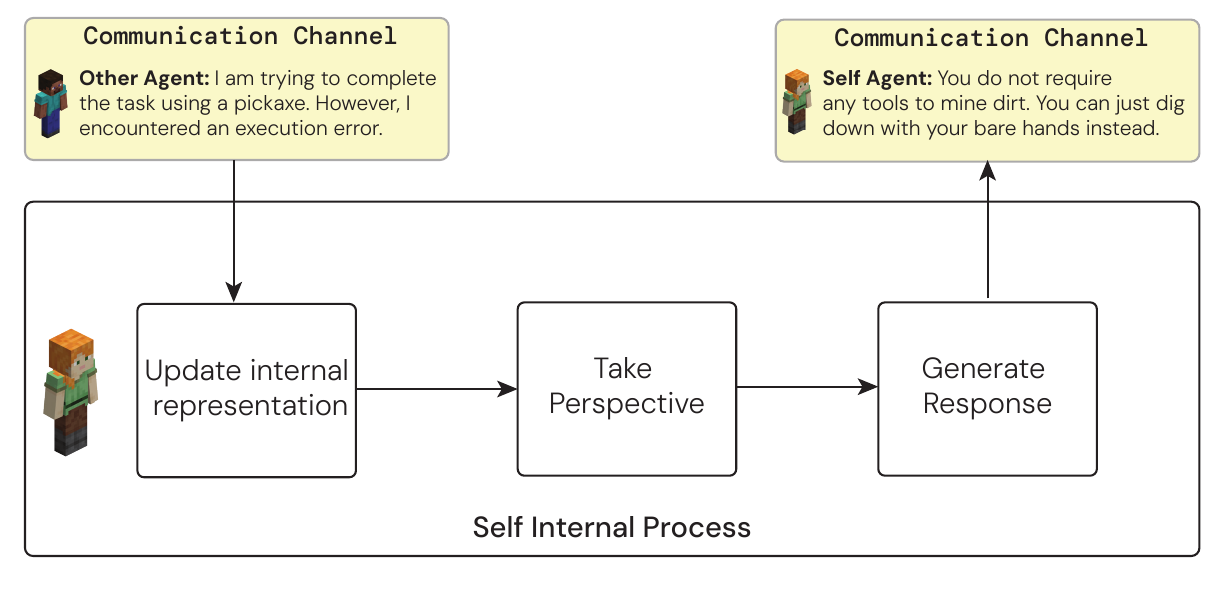}
\caption{Flow of \collabvoyager communication components used to generate a natural language response. The agents' internal process consists of two stages: (i) internal representation update and (ii) perspective-taking. Leveraging both processes the agent can then generate a natural language response and send it through the Minecraft chat.}
\label{fig:flow}
\end{figure}

Perspective-taking represents a central part of our framework since it allows \collabvoyager agents to take advantage of the internal causal structure of other agents in order to aid more efficient and grounded communication. In practice, perspective-taking corresponds to an LLM call where the prompt contains instructions and relevant information required to understand the internal state of other \collabvoyager agents during communication. \Cref{fig:flow} showcases how perspective-taking integrates with the overall communication mechanism. 

Once the agent receives a message in the Minecraft chat, a first LLM call is executed where the agent updates the stateful internal representation about the agent that sent the message. This update process is shown in \Cref{fig:beliefs-evolution-rounds}. Subsequently, the updated internal representation is used as context together with the rest of the conversation to take perspective of the situation of the other \collabvoyager agent. Similar to the first stage, perspective-taking is implemented as a LLM call, with the following prompt:

\vspace{0.25cm}
\begin{tcolorbox}[title=Prompt for perspective-taking, label=box:prompt-perspective]
You are a Minecraft agent named \{\{name\}\} and you are having a conversation with another agent named \{\{other\_name\}\}. \\

Based on the current conversation and your knowledge about the other agent, \{\{other\_name\}\}, take the other agent's perspective to assess and describe your current understanding, knowledge state, and likely needs from \{\{other\_name\}\}'s perspective. \\

Here is the current conversation between you and \{\{other\_name\}\}:\{conversation\} \\

Here is your mental model of \{\{other\_name\}\}: \{world\_model\} \\

Perspective Analysis: 
\end{tcolorbox}

Below you can find an example of perspective-taking from Minecraft gameplay:

\begin{tcolorbox}[title=Expert agent taking perspective of weak agent, label=box:pers-prompt]
Weak’s current knowledge state includes understanding the biome they are in, the time of day, and the fact that there are
dark oak logs nearby. Weak knows that a wooden axe is necessary for mining logs, and that crafting this tool will be a
required step before proceeding with the task. Weak’s likely needs from Strong include further assistance in navigating the environment during the night to find a tree
or waiting until the day.
\end{tcolorbox}

\subsection{Episodic Memory}

We implement episodic memory as a Retrieval-Augmented Generation (RAG) using LangChain \citep{langchain2024}. Specifically, we embed all the episodes where the agent failed to complete the task using \textbf{text-embedding-ada-002}. An episode consists of the context used to generate the action, the code itself and the corresponding critic message. Subsequently, when the agent generates a new action, the RAG is queried to retrieve the \textbf{k=5} most relevant failure episodes with respect to the task at hand. Lastly, to ensure we do not populate the context window with too many tokens, we generate a summary (using an LLM) of the retrieved episodes and feed this summary in the context used to generate the action. Below you can find the corresponding prompts for generating the episodic summary:

\begin{tcolorbox}[title=System message, label=box:sys-episodic]
You are a helpful assistant tasked with summarizing past experience episodes and pointing out the causes of failure. Create a concise summary.
\end{tcolorbox}

\begin{tcolorbox}[title=Prompt message, label=box:prompt-episodic]
Please summarize these episodes and why they failed:\{combined\_episodes\}
\end{tcolorbox}

\subsection{Procedural Memory}

The procedural memory has an identical design and functionality with the skill library introduced in Voyager \citep{wang2023voyager}. Once a task gets successfully completed, we store the corresponding piece of code such that we can later reference it in adjacent tasks. This becomes increasingly important in tasks that require multiple steps, like crafting a pickaxe: the agent needs to first create a crafting table, create wooden sticks and only then attempt to craft the pickaxe. When we run the \collabvoyager framework in an open-ended setting, the procedural memory allows the agent to continuously evolve and create an increasingly large collection of reusable skills.

\subsection{Semantic Memory}

As described in \Cref{ssec:memory-subsys}, semantic memory stores certain beliefs the agent has about the environment, coming from experience with the environment and the other \collabvoyager agents. Whenever \collabvoyager attempts a new task, it creates a certain belief about how the task should be solved. For example, the agent might believe it requires tools to mine a block of dirt. This belief is then added to the context of the LLM when the environment action is generated. As highlighted in \Cref{sec:intro}, one of the primary failure cases of Voyager are the false beliefs that come from this LLM generation. However, compared to Voyager \citep{wang2023voyager}, we leverage the communication with other \collabvoyager agents such that these beliefs get updated and refined. Below we showcase an example of false belief correction when trying to collect a block of wood:

\begin{tcolorbox}[title=Initial belief inside the semantic memory, label=box:response-semantic-1]
\{"How to mine 1 wood log in Minecraft?": "Answer: To mine 1 wood log in Minecraft, you need to use an axe."\}
\end{tcolorbox}

\begin{tcolorbox}[title=Updated belief after communication, label=box:response-semantic-1]
\{"How to mine 1 wood log in Minecraft?": "Answer: To mine 1 wood log in Minecraft, you need to punch a tree with your bare hands."\}
\end{tcolorbox}
In addition to task-related memory, the \collabvoyager agent maintains a stateful set of beliefs about the other agents in the environment that are constantly updated across episodes. An example of this is presented in \Cref{fig:beliefs-evolution-rounds}, where the agent updates its representation based on the contents of the conversation. Thus, keeping track of various information and facts about other \collabvoyager agents acts as a semantic memory retrieval when agents communicate.

\section{Sample Conversation}
\label{appendix:sample-conv}

Each communication round between two \collabvoyager agents comprises of six sending messages in a turn-based manner through the Minecraft build-in chat. Below you can find a sample conversation showcasing how the expert agent (Strong) helps the weak (Weak) agent. 

\begin{tcolorbox}[title=Conversation about completing the task: "Mine 1 wood log"]

\textbf{Weak}: Hey, can you help me with Mine 1 wood log? \\

\textbf{Strong}: Absolutely, I can help you with that. Given your current location in the dark forest biome and the presence of dark oak logs around you, you should be able to mine a log without any issues. Here's the code you can use: `mine('dark\_oak\_log')`. This command will make your agent mine a dark oak log. Just make sure you are close enough to the tree when executing this. \\

\textbf{Weak}:  Thank you for the helpful guidance, Strong! However, in my case, I am currently in a dark forest biome as well, but I am having trouble finding a tree during the night. Could you suggest a way to find a tree during the day so I can mine the wood log? I also noticed that I don't have a wooden axe in my inventory, so I may need to craft one before attempting to mine the log. I appreciate your assistance in this matter. \\

\textbf{Strong}: Of course, I understand your predicament. Since it's currently night time, visibility might be low which makes it challenging to locate a tree. You can wait until daybreak for better visibility. Use the `wait\_until\_day()` function to wait until it's daytime. For crafting a wooden axe, you would need 3 wood planks and 2 sticks. You can obtain wood planks from a wood log by using the `craft('wooden\_plank', wood\_log)` command and sticks from 2 wooden planks using the `craft('stick', wooden\_plank)` command. Once you have these, you can craft a wooden axe using the `craft('wooden\_axe', wooden\_plank, stick)` function. Here's the code:

\begin{verbatim}
```
wait_until_day()
craft(`wooden_plank', wood_log)
craft(`stick', wooden_plank)
craft(`wooden_axe', wooden_plank, stick)
```
\end{verbatim}

Once you have your axe and it's day time, you should be able to easily locate and mine a tree for the wood log. \\ 

\textbf{Weak}:  Thank you, Strong. I appreciate your help.

\end{tcolorbox}

\section{Fine-tuning setup}
\label{appendix:fine}
\subsection{Execution Logs}
For the first fine-tuning variant, we use execution logs generated by a Voyager agent powered by GPT-4. Specifically, these logs contain input and output text information obtained from the components specific to the architecture, such as the actor or the critic. We find this to be a meaningful way to transfer knowledge as well as reasoning capabilities from GPT-4 to Mistral-7B. To perform the fine-tuning, we use LoRA \citep{hu2021lora} due to its low computational requirements. 

\subsection{Wiki Data and Documentation}

The second fine-tuning variation we consider builds upon the initial setup presented above and incorporates more structured information about Minecraft by including the Minecraft Wiki \citep{fan2022minedojo} and Minecraft API documentation. Similar to the first variant we use LoRA \citep{hu2021lora}.

\section{Prompts}
\label{appendix:prompts}

\begin{tcolorbox}[title=Prompt for updating partner beliefs, label=box:response-semantic-1]
You are a Minecraft agent. \\ 

You just had a conversation with another agent based on a task you are trying to solve. \\

Based on the contents of the conversation and the previous beliefs, you have to create a set of beliefs that represent your perception of the other agent.
\end{tcolorbox}

\begin{tcolorbox}[title=Prompt for creating interaction beliefs, label=box:response-semantic-1]
You are a Minecraft agent. \\

You just had a conversation with another agent based on a task you are trying to solve. \\

Based on the contents of the conversation and the previous beliefs, you have to create a set of beliefs that that can help you complete the task.
\end{tcolorbox}

\vspace{0.25cm}
\begin{tcolorbox}[title=Prompt for perspective-taking, label=box:prompt-perspective]
You are a Minecraft agent named \{\{name\}\} and you are having a conversation with another agent named \{\{other\_name\}\}. \\

Based on the current conversation and your knowledge about the other agent, \{\{other\_name\}\}, take the other agent's perspective to assess and describe your current understanding, knowledge state, and likely needs from \{\{other\_name\}\}'s perspective. \\

Here is the current conversation between you and \{\{other\_name\}\}:\{conversation\} \\

Here is your mental model of \{\{other\_name\}\}: \{world\_model\} \\

Perspective Analysis: 
\end{tcolorbox}

\section{LLMs} \label{appendix:llm}
This section details the technical specifications of the Large Language Models (LLMs) and related infrastructure used to implement the \collabvoyager framework and conduct the experiments presented in this paper.

All LLM calls, unless explicitly stated otherwise, were executed via API calls to external services. The exception to this is the fine-tuned model, which was ran and trained locally.

The specific models and their corresponding access methods are outlined below:

\begin{itemize}
    \item \textbf{Llama 3.1-8B-Instruct}: This model was accessed through the Lambda Inference API.\footnote{\url{https://docs.lambdalabs.com/public-cloud/lambda-inference-api/}} This API provided a reliable and efficient interface for interacting with the Llama 3.1-8B-Instruct model, enabling seamless integration within the \collabvoyager framework.
    \item \textbf{GPT-4}:  Accessed via the OpenAI API. The OpenAI API offered access to the advanced capabilities of the GPT-4 model, crucial for establishing performance baselines and for the expert agent in collaborative settings.
    \item \textbf{Mistral-7B Instruct v0.2, Mixtral-8x7B Instruct and Llama 3.1-70B}: These Mistral models were accessed through the Together AI API.\footnote{\url{https://api.together.ai/models}} The Together AI API provided a platform for utilizing these open-weight models, allowing for a comparative analysis within the Voyager and \collabvoyager frameworks.
\end{itemize}

%% file: references.bib
@inproceedings{zhao2024expel,
  title={Expel: Llm agents are experiential learners},
  author={Zhao, Andrew and Huang, Daniel and Xu, Quentin and Lin, Matthieu and Liu, Yong-Jin and Huang, Gao},
  booktitle={Proceedings of the AAAI Conference on Artificial Intelligence},
  volume={38},
  number={17},
  pages={19632--19642},
  year={2024}
}

@article{wang2023describe,
  title={Describe, explain, plan and select: Interactive planning with large language models enables open-world multi-task agents},
  author={Wang, Zihao and Cai, Shaofei and Chen, Guanzhou and Liu, Anji and Ma, Xiaojian and Liang, Yitao},
  journal={arXiv preprint arXiv:2302.01560},
  year={2023}
}

@article{nottingham2024skill,
  title={Skill set optimization: Reinforcing language model behavior via transferable skills},
  author={Nottingham, Kolby and Majumder, Bodhisattwa Prasad and Mishra, Bhavana Dalvi and Singh, Sameer and Clark, Peter and Fox, Roy},
  journal={arXiv preprint arXiv:2402.03244},
  year={2024}
}

@article{prasad2023adapt,
  title={Adapt: As-needed decomposition and planning with language models},
  author={Prasad, Archiki and Koller, Alexander and Hartmann, Mareike and Clark, Peter and Sabharwal, Ashish and Bansal, Mohit and Khot, Tushar},
  journal={arXiv preprint arXiv:2311.05772},
  year={2023}
}

@article{majumder2023clin,
  title={Clin: A continually learning language agent for rapid task adaptation and generalization},
  author={Majumder, Bodhisattwa Prasad and Mishra, Bhavana Dalvi and Jansen, Peter and Tafjord, Oyvind and Tandon, Niket and Zhang, Li and Callison-Burch, Chris and Clark, Peter},
  journal={arXiv preprint arXiv:2310.10134},
  year={2023}
}

@article{yu2024adam,
	title        = {ADAM: An Embodied Causal Agent in Open-World Environments},
	author       = {Yu, Shu and Lu, Chaochao},
	year         = 2024,
	journal      = {arXiv preprint arXiv:2410.22194}
}

@article{li2024optimus,
	title        = {Optimus-1: Hybrid multimodal memory empowered agents excel in long-horizon tasks},
	author       = {Li, Zaijing and Xie, Yuquan and Shao, Rui and Chen, Gongwei and Jiang, Dongmei and Nie, Liqiang},
	year         = 2024,
	journal      = {arXiv preprint arXiv:2408.03615}
}

@misc{mineflayer,
	title        = {Mineflayer},
	author       = {PrismarineJS},
	year         = 2021,
	url          = {https://prismarinejs.github.io/mineflayer/}
}

@misc{langchain2024,
	title        = {LangChain documentation},
	author       = {LangChain},
	year         = 2024,
	url          = {https://python.langchain.com/},
	note         = {Accessed: 2024-11-19}
}

@article{das2024larimar,
	title        = {Larimar: Large Language Models with Episodic Memory Control},
	author       = {Das, Payel and Chaudhury, Subhajit and Nelson, Elliot and Melnyk, Igor and Swaminathan, Sarath and Dai, Sihui and Lozano, Aur{\'e}lie and Kollias, Georgios and Chenthamarakshan, Vijil and Dan, Soham and others},
	year         = 2024,
	journal      = {arXiv preprint arXiv:2403.11901}
}

@article{bornstein2017reminders,
	title        = {Reminders of past choices bias decisions for reward in humans},
	author       = {Bornstein, Aaron M and Khaw, Mel W and Shohamy, Daphna and Daw, Nathaniel D},
	year         = 2017,
	journal      = {Nature Communications},
	publisher    = {Nature Publishing Group UK London},
	volume       = 8,
	number       = 1,
	pages        = 15958
}

@article{kosinski2023theory,
	title        = {Theory of mind might have spontaneously emerged in large language models},
	author       = {Kosinski, Michal},
	year         = 2023,
	journal      = {arXiv preprint arXiv:2302.02083}
}

@article{ahn2022can,
	title        = {Do as i can, not as i say: Grounding language in robotic affordances},
	author       = {Ahn, Michael and Brohan, Anthony and Brown, Noah and Chebotar, Yevgen and Cortes, Omar and David, Byron and Finn, Chelsea and Fu, Chuyuan and Gopalakrishnan, Keerthana and Hausman, Karol and others},
	year         = 2022,
	journal      = {arXiv preprint arXiv:2204.01691}
}

@article{lin2023text2motion,
	title        = {Text2motion: From natural language instructions to feasible plans},
	author       = {Lin, Kevin and Agia, Christopher and Migimatsu, Toki and Pavone, Marco and Bohg, Jeannette},
	year         = 2023,
	journal      = {Autonomous Robots},
	publisher    = {Springer},
	volume       = 47,
	number       = 8,
	pages        = {1345--1365}
}

@inproceedings{khattab2024dspy,
	title        = {DSPy: Compiling Declarative Language Model Calls into Self-Improving Pipelines},
	author       = {Khattab, Omar and Singhvi, Arnav and Maheshwari, Paridhi and Zhang, Zhiyuan and Santhanam, Keshav and Vardhamanan, Sri and Haq, Saiful and Sharma, Ashutosh and Joshi, Thomas T. and Moazam, Hanna and Miller, Heather and Zaharia, Matei and Potts, Christopher},
	year         = 2024,
	journal      = {The Twelfth International Conference on Learning Representations}
}

@article{khattab2022demonstrate,
	title        = {Demonstrate-Search-Predict: Composing Retrieval and Language Models for Knowledge-Intensive {NLP}},
	author       = {Khattab, Omar and Santhanam, Keshav and Li, Xiang Lisa and Hall, David and Liang, Percy and Potts, Christopher and Zaharia, Matei},
	year         = 2022,
	journal      = {arXiv preprint arXiv:2212.14024}
}

@article{dubey2024llama,
	title        = {The llama 3 herd of models},
	author       = {Dubey, Abhimanyu and Jauhri, Abhinav and Pandey, Abhinav and Kadian, Abhishek and Al-Dahle, Ahmad and Letman, Aiesha and Mathur, Akhil and Schelten, Alan and Yang, Amy and Fan, Angela and others},
	year         = 2024,
	journal      = {arXiv preprint arXiv:2407.21783}
}

@article{du2023improving,
	title        = {Improving factuality and reasoning in language models through multiagent debate},
	author       = {Du, Yilun and Li, Shuang and Torralba, Antonio and Tenenbaum, Joshua B and Mordatch, Igor},
	year         = 2023,
	journal      = {arXiv preprint arXiv:2305.14325}
}

@article{zhang2023building,
	title        = {Building cooperative embodied agents modularly with large language models},
	author       = {Zhang, Hongxin and Du, Weihua and Shan, Jiaming and Zhou, Qinhong and Du, Yilun and Tenenbaum, Joshua B and Shu, Tianmin and Gan, Chuang},
	year         = 2023,
	journal      = {arXiv preprint arXiv:2307.02485}
}

@article{bubeck2023sparks,
	title        = {Sparks of artificial general intelligence: Early experiments with gpt-4},
	author       = {Bubeck, S{\'e}bastien and Chandrasekaran, Varun and Eldan, Ronen and Gehrke, Johannes and Horvitz, Eric and Kamar, Ece and Lee, Peter and Lee, Yin Tat and Li, Yuanzhi and Lundberg, Scott and others},
	year         = 2023,
	journal      = {arXiv preprint arXiv:2303.12712}
}

@article{wilf2023think,
	title        = {Think Twice: Perspective-Taking Improves Large Language Models' Theory-of-Mind Capabilities},
	author       = {Wilf, Alex and Lee, Sihyun Shawn and Liang, Paul Pu and Morency, Louis-Philippe},
	year         = 2023,
	journal      = {arXiv preprint arXiv:2311.10227}
}

@article{jin2024mmtom,
	title        = {Mmtom-qa: Multimodal theory of mind question answering},
	author       = {Jin, Chuanyang and Wu, Yutong and Cao, Jing and Xiang, Jiannan and Kuo, Yen-Ling and Hu, Zhiting and Ullman, Tomer and Torralba, Antonio and Tenenbaum, Joshua B and Shu, Tianmin},
	year         = 2024,
	journal      = {arXiv preprint arXiv:2401.08743}
}

@article{li2023theory,
	title        = {Theory of mind for multi-agent collaboration via large language models},
	author       = {Li, Huao and Chong, Yu Quan and Stepputtis, Simon and Campbell, Joseph and Hughes, Dana and Lewis, Michael and Sycara, Katia},
	year         = 2023,
	journal      = {arXiv preprint arXiv:2310.10701}
}

@article{zhou2023far,
	title        = {How FaR Are Large Language Models From Agents with Theory-of-Mind?},
	author       = {Zhou, Pei and Madaan, Aman and Potharaju, Srividya Pranavi and Gupta, Aditya and McKee, Kevin R and Holtzman, Ari and Pujara, Jay and Ren, Xiang and Mishra, Swaroop and Nematzadeh, Aida and others},
	year         = 2023,
	journal      = {arXiv preprint arXiv:2310.03051}
}

@article{sclar2023minding,
	title        = {Minding language models'(lack of) theory of mind: A plug-and-play multi-character belief tracker},
	author       = {Sclar, Melanie and Kumar, Sachin and West, Peter and Suhr, Alane and Choi, Yejin and Tsvetkov, Yulia},
	year         = 2023,
	journal      = {arXiv preprint arXiv:2306.00924}
}

@article{hao2023reasoning,
	title        = {Reasoning with language model is planning with world model},
	author       = {Hao, Shibo and Gu, Yi and Ma, Haodi and Hong, Joshua Jiahua and Wang, Zhen and Wang, Daisy Zhe and Hu, Zhiting},
	year         = 2023,
	journal      = {arXiv preprint arXiv:2305.14992}
}

@article{ullman2023large,
	title        = {Large language models fail on trivial alterations to theory-of-mind tasks},
	author       = {Ullman, Tomer},
	year         = 2023,
	journal      = {arXiv preprint arXiv:2302.08399}
}

@inproceedings{rabinowitz2018machine,
	title        = {Machine theory of mind},
	author       = {Rabinowitz, Neil and Perbet, Frank and Song, Francis and Zhang, Chiyuan and Eslami, SM Ali and Botvinick, Matthew},
	year         = 2018,
	booktitle    = {International conference on machine learning},
	pages        = {4218--4227},
	organization = {PMLR}
}

@inproceedings{chuang2020using,
	title        = {Using machine theory of mind to learn agent social network structures from observed interactive behaviors with targets},
	author       = {Chuang, Yun-Shiuan and Hung, Hsin-Yi and Gamborino, Edwinn and Goh, Joshua Oon Soo and Huang, Tsung-Ren and Chang, Yu-Ling and Yeh, Su-Ling and Fu, Li-Chen},
	year         = 2020,
	booktitle    = {2020 29th IEEE International Conference on Robot and Human Interactive Communication (RO-MAN)},
	pages        = {1013--1019},
	organization = {IEEE}
}

@inproceedings{baker2011bayesian,
	title        = {Bayesian theory of mind: Modeling joint belief-desire attribution},
	author       = {Baker, Chris and Saxe, Rebecca and Tenenbaum, Joshua},
	year         = 2011,
	booktitle    = {Proceedings of the annual meeting of the cognitive science society},
	volume       = 33,
	number       = 33
}

@inproceedings{grassiotto2021cogtom,
	title        = {CogToM: A Cognitive Architecture Implementation of the Theory of Mind.},
	author       = {Grassiotto, Fabio and Costa, Paula Dornhofer Paro},
	year         = 2021,
	booktitle    = {ICAART (2)},
	pages        = {546--553}
}

@inproceedings{shu2021agent,
	title        = {Agent: A benchmark for core psychological reasoning},
	author       = {Shu, Tianmin and Bhandwaldar, Abhishek and Gan, Chuang and Smith, Kevin and Liu, Shari and Gutfreund, Dan and Spelke, Elizabeth and Tenenbaum, Joshua and Ullman, Tomer},
	year         = 2021,
	booktitle    = {International conference on machine learning},
	pages        = {9614--9625},
	organization = {PMLR}
}

@article{wang2023voyager,
	title        = {Voyager: An open-ended embodied agent with large language models},
	author       = {Wang, Guanzhi and Xie, Yuqi and Jiang, Yunfan and Mandlekar, Ajay and Xiao, Chaowei and Zhu, Yuke and Fan, Linxi and Anandkumar, Anima},
	year         = 2023,
	journal      = {arXiv preprint arXiv:2305.16291}
}

@article{gandhi2024understanding,
	title        = {Understanding social reasoning in language models with language models},
	author       = {Gandhi, Kanishk and Fr{\"a}nken, Jan-Philipp and Gerstenberg, Tobias and Goodman, Noah},
	year         = 2024,
	journal      = {Advances in Neural Information Processing Systems},
	volume       = 36
}

@article{vezhnevets2023generative,
	title        = {Generative agent-based modeling with actions grounded in physical, social, or digital space using Concordia},
	author       = {Vezhnevets, Alexander Sasha and Agapiou, John P and Aharon, Avia and Ziv, Ron and Matyas, Jayd and Du{\'e}{\~n}ez-Guzm{\'a}n, Edgar A and Cunningham, William A and Osindero, Simon and Karmon, Danny and Leibo, Joel Z},
	year         = 2023,
	journal      = {arXiv preprint arXiv:2312.03664}
}

@article{zhang2024combo,
	title        = {COMBO: Compositional World Models for Embodied Multi-Agent Cooperation},
	author       = {Zhang, Hongxin and Wang, Zeyuan and Lyu, Qiushi and Zhang, Zheyuan and Chen, Sunli and Shu, Tianmin and Du, Yilun and Gan, Chuang},
	year         = 2024,
	journal      = {arXiv preprint arXiv:2404.10775}
}

@article{talebirad2023multi,
	title        = {Multi-agent collaboration: Harnessing the power of intelligent llm agents},
	author       = {Talebirad, Yashar and Nadiri, Amirhossein},
	year         = 2023,
	journal      = {arXiv preprint arXiv:2306.03314}
}

@article{hinton2015distilling,
	title        = {Distilling the Knowledge in a Neural Network},
	author       = {Hinton, Geoffrey},
	year         = 2015,
	journal      = {arXiv preprint arXiv:1503.02531}
}

@article{jiang2024mixtral,
	title        = {Mixtral of experts},
	author       = {Jiang, Albert Q and Sablayrolles, Alexandre and Roux, Antoine and Mensch, Arthur and Savary, Blanche and Bamford, Chris and Chaplot, Devendra Singh and Casas, Diego de las and Hanna, Emma Bou and Bressand, Florian and others},
	year         = 2024,
	journal      = {arXiv preprint arXiv:2401.04088}
}

@misc{jiang2023mistral7b,
	title        = {Mistral 7B},
	author       = {Albert Q. Jiang and Alexandre Sablayrolles and Arthur Mensch and Chris Bamford and Devendra Singh Chaplot and Diego de las Casas and Florian Bressand and Gianna Lengyel and Guillaume Lample and Lucile Saulnier and Lélio Renard Lavaud and Marie-Anne Lachaux and Pierre Stock and Teven Le Scao and Thibaut Lavril and Thomas Wang and Timothée Lacroix and William El Sayed},
	year         = 2023,
	url          = {https://arxiv.org/abs/2310.06825},
	eprint       = {2310.06825},
	archiveprefix = {arXiv},
	primaryclass = {cs.CL}
}

@inproceedings{rao1995bdi,
	title        = {BDI agents: from theory to practice.},
	author       = {Rao, Anand S and Georgeff, Michael P and others},
	year         = 1995,
	booktitle    = {Icmas},
	volume       = 95,
	pages        = {312--319}
}

@article{hu2021lora,
	title        = {Lora: Low-rank adaptation of large language models},
	author       = {Hu, Edward J and Shen, Yelong and Wallis, Phillip and Allen-Zhu, Zeyuan and Li, Yuanzhi and Wang, Shean and Wang, Lu and Chen, Weizhu},
	year         = 2021,
	journal      = {arXiv preprint arXiv:2106.09685}
}

@article{zare2024survey,
	title        = {A survey of imitation learning: Algorithms, recent developments, and challenges},
	author       = {Zare, Maryam and Kebria, Parham M and Khosravi, Abbas and Nahavandi, Saeid},
	year         = 2024,
	journal      = {IEEE Transactions on Cybernetics},
	publisher    = {IEEE}
}

@article{schulman2017proximal,
	title        = {Proximal policy optimization algorithms},
	author       = {Schulman, John and Wolski, Filip and Dhariwal, Prafulla and Radford, Alec and Klimov, Oleg},
	year         = 2017,
	journal      = {arXiv preprint arXiv:1707.06347}
}

@article{hafner2023mastering,
	title        = {Mastering diverse domains through world models},
	author       = {Hafner, Danijar and Pasukonis, Jurgis and Ba, Jimmy and Lillicrap, Timothy},
	year         = 2023,
	journal      = {arXiv preprint arXiv:2301.04104}
}

@inproceedings{cobbe2019quantifying,
	title        = {Quantifying generalization in reinforcement learning},
	author       = {Cobbe, Karl and Klimov, Oleg and Hesse, Chris and Kim, Taehoon and Schulman, John},
	year         = 2019,
	booktitle    = {International conference on machine learning},
	pages        = {1282--1289},
	organization = {PMLR}
}

@article{fan2022minedojo,
	title        = {Minedojo: Building open-ended embodied agents with internet-scale knowledge},
	author       = {Fan, Linxi and Wang, Guanzhi and Jiang, Yunfan and Mandlekar, Ajay and Yang, Yuncong and Zhu, Haoyi and Tang, Andrew and Huang, De-An and Zhu, Yuke and Anandkumar, Anima},
	year         = 2022,
	journal      = {Advances in Neural Information Processing Systems},
	volume       = 35,
	pages        = {18343--18362}
}

@article{achiam2023gpt,
	title        = {Gpt-4 technical report},
	author       = {Achiam, Josh and Adler, Steven and Agarwal, Sandhini and Ahmad, Lama and Akkaya, Ilge and Aleman, Florencia Leoni and Almeida, Diogo and Altenschmidt, Janko and Altman, Sam and Anadkat, Shyamal and others},
	year         = 2023,
	journal      = {arXiv preprint arXiv:2303.08774}
}

@article{kolve2017ai2,
	title        = {AI2-THOR: An Interactive 3D Environment for Visual AI},
	author       = {Kolve, Eric and Mottaghi, Roozbeh and Gordon, Daniel and Zhu, Yuke and Gupta, Abhinav and Farhadi, Ali},
	year         = 2017,
	journal      = {arXiv preprint arXiv:1712.05474}
}

@inproceedings{savva2019habitat,
	title        = {Habitat: A Platform for Embodied AI Research},
	author       = {Savva, Manolis and Kadian, Abhishek and Maksymets, Oleksandr and Zhao, Yili and Wijmans, Erik and Jain, Bhavana and Straub, Julian and Liu, Jia and Koltun, Vladlen and Malik, Jitendra and others},
	year         = 2019,
	booktitle    = {Proceedings of the IEEE/CVF International Conference on Computer Vision},
	pages        = {9339--9347}
}

@article{han2024parameter,
	title        = {Parameter-efficient fine-tuning for large models: A comprehensive survey},
	author       = {Han, Zeyu and Gao, Chao and Liu, Jinyang and Zhang, Sai Qian and others},
	year         = 2024,
	journal      = {arXiv preprint arXiv:2403.14608}
}

@article{puig2018virtualhome,
	title        = {VirtualHome: Simulating household activities via programs},
	author       = {Puig, Xavier and Ra, Kevin and Boben, Marko and Li, Jiaman and Wang, Tingwu and Fidler, Sanja and Torralba, Antonio},
	year         = 2018,
	journal      = {Proceedings of the IEEE Conference on Computer Vision and Pattern Recognition},
	pages        = {8494--8502}
}

@inproceedings{shridhar2020alfred,
	title        = {ALFRED: A Benchmark for Interpreting Grounded Instructions for Everyday Tasks},
	author       = {Shridhar, Mohit and Thomason, Jesse and Gordon, Daniel and Bisk, Yonatan and Han, Winson and Mottaghi, Roozbeh and Zettlemoyer, Luke and Fox, Dieter},
	year         = 2020,
	booktitle    = {Proceedings of the IEEE/CVF Conference on Computer Vision and Pattern Recognition},
	pages        = {10740--10749}
}

@article{parisi2019continual,
	title        = {Continual lifelong learning with neural networks: A review},
	author       = {Parisi, German I and Kemker, Ronald and Part, Jose L and Kanan, Christopher and Wermter, Stefan},
	year         = 2019,
	journal      = {Neural Networks},
	publisher    = {Elsevier},
	volume       = 113,
	pages        = {54--71}
}

@article{wang2023comprehensive,
	title        = {A comprehensive survey of continual learning: Theory, method and application},
	author       = {Wang, Liyuan and Zhang, Xingxing and Su, Hang and Zhu, Jun},
	year         = 2023,
	journal      = {arXiv preprint arXiv:2302.00487}
}

@article{humphrey1976social,
	title        = {The social function of intellect},
	author       = {Humphrey, Nicholas K},
	year         = 1976,
	journal      = {Growing points in ethology},
	publisher    = {Cambridge University Press},
	pages        = {303--317}
}

@article{dunbar1998social,
	title        = {The social brain hypothesis},
	author       = {Dunbar, Robin IM},
	year         = 1998,
	journal      = {Evolutionary Anthropology: Issues, News, and Reviews},
	publisher    = {Wiley Online Library},
	volume       = 6,
	number       = 5,
	pages        = {178--190}
}

@article{tomasello1993cultural,
	title        = {Cultural learning},
	author       = {Tomasello, Michael and Kruger, Ann Cale and Ratner, Hilary Horn},
	year         = 1993,
	journal      = {Behavioral and brain sciences},
	publisher    = {Cambridge University Press},
	volume       = 16,
	number       = 3,
	pages        = {495--511}
}

@article{devaine2014social,
	title        = {The social Bayesian brain: does mentalizing make a difference when we learn?},
	author       = {Devaine, Marie and Hollard, Guillaume and Daunizeau, Jean},
	year         = 2014,
	journal      = {PLoS computational biology},
	publisher    = {Public Library of Science San Francisco, USA},
	volume       = 10,
	number       = 12,
	pages        = {e1003992}
}

@article{tomasello2016cultural,
	title        = {Cultural learning redux},
	author       = {Tomasello, Michael},
	year         = 2016,
	journal      = {Child development},
	publisher    = {Wiley Online Library},
	volume       = 87,
	number       = 3,
	pages        = {643--653}
}

@article{premack1978does,
	title        = {Does the chimpanzee have a theory of mind?},
	author       = {Premack, David and Woodruff, Guy},
	year         = 1978,
	journal      = {Behavioral and brain sciences},
	publisher    = {Cambridge University Press},
	volume       = 1,
	number       = 4,
	pages        = {515--526}
}

@incollection{baron1999evolution,
  title     = "The evolution of a theory of mind",
  author    = "Baron-Cohen, Simon",
  editor    = "Corballis, Michael C",
  booktitle = "The descent of mind: Psychological perspectives on hominid
               evolution (pp",
  publisher = "Oxford University Press, xii",
  volume    =  361,
  pages     = "261--277",
  year      =  1999,
  url       = "https://psycnet.apa.org/record/1999-04113-013"
}

@book{laird2019soar,
	title        = {The Soar cognitive architecture},
	author       = {Laird, John E},
	year         = 2019,
	publisher    = {MIT press}
}

@article{Wellman1990-qh,
	title        = {From simple desires to ordinary beliefs: the early development of everyday psychology},
	author       = {Wellman, H M and Woolley, J D},
	year         = 1990,
	month        = {1~} # {jun},
	journal      = {Cognition},
	publisher    = {Elsevier},
	volume       = 35,
	number       = 3,
	pages        = {245--275},
	doi          = {10.1016/0010-0277(90)90024-e},
	url          = {http://dx.doi.org/10.1016/0010-0277(90)90024-E},
	language     = {en}
}

@book{Perner-book,
	title        = {{Understanding the Representational Mind}},
	author       = {Perner, Josef},
	year         = 1993,
	month        = {03},
	publisher    = {The MIT Press},
	doi          = {10.7551/mitpress/6988.001.0001},
	isbn         = 9780262316033,
	url          = {https://doi.org/10.7551/mitpress/6988.001.0001}
}

@article{Ochsner2004-jd,
	title        = {For better or for worse: neural systems supporting the cognitive down- and up-regulation of negative emotion},
	author       = {Ochsner, Kevin N and Ray, Rebecca D and Cooper, Jeffrey C and Robertson, Elaine R and Chopra, Sita and Gabrieli, John D E and Gross, James J},
	year         = 2004,
	month        = {1~} # {oct},
	journal      = {Neuroimage},
	publisher    = {Elsevier BV},
	volume       = 23,
	number       = 2,
	pages        = {483--499},
	doi          = {10.1016/j.neuroimage.2004.06.030},
	url          = {http://dx.doi.org/10.1016/j.neuroimage.2004.06.030},
	language     = {en}
}

@article{Saxe2003-zr,
	title        = {People thinking about thinking people. The role of the temporo-parietal junction in ``theory of mind''},
	author       = {Saxe, R and Kanwisher, N},
	year         = 2003,
	month        = aug,
	journal      = {Neuroimage},
	publisher    = {Neuroimage},
	volume       = 19,
	number       = 4,
	pages        = {1835--1842},
	doi          = {10.1016/s1053-8119(03)00230-1},
	url          = {https://pubmed.ncbi.nlm.nih.gov/12948738/},
	language     = {en}
}

@article{Klein2009-vz,
	title        = {Evolution and episodic memory: An analysis and demonstration of a social function of episodic recollection},
	author       = {Klein, S and Cosmides, L and Gangi, C and Jackson, Betsy and Tooby, J and Costabile, Kristi A},
	year         = 2009,
	month        = {21~} # {apr},
	journal      = {Soc. Cogn.},
	volume       = 27,
	number       = 2,
	pages        = {283--319},
	doi          = {10.1521/SOCO.2009.27.2.283},
	url          = {http://guilfordjournals.com/doi/10.1521/soco.2009.27.2.283}
}

@article{Klein1993-vu,
	title        = {Self-knowledge of an amnesic patient: Toward a neuropsychology of personality and social psychology},
	author       = {Klein, Stanley B and Loftus, Judith and Kihlstrom, John F},
	year         = 1993,
	journal      = {b; Tulving},
	volume       = 1983,
	pages        = 1989
}

@article{Binder2011-gq,
	title        = {The neurobiology of semantic memory},
	author       = {Binder, Jeffrey R and Desai, Rutvik H},
	year         = 2011,
	month        = {1~} # {nov},
	journal      = {Trends Cogn. Sci.},
	publisher    = {Elsevier BV},
	volume       = 15,
	number       = 11,
	pages        = {527--536},
	doi          = {10.1016/j.tics.2011.10.001},
	url          = {http://dx.doi.org/10.1016/j.tics.2011.10.001},
	language     = {en}
}

@article{austen1996information,
	title        = {Information aggregation, rationality, and the Condorcet jury theorem},
	author       = {Austen-Smith, David and Banks, Jeffrey S},
	year         = 1996,
	journal      = {American political science review},
	publisher    = {Cambridge University Press},
	volume       = 90,
	number       = 1,
	pages        = {34--45}
}

@article{white2025collaborating,
	title        = {Collaborating Action by Action: A Multi-agent LLM Framework for Embodied Reasoning},
	author       = {White, Isadora and Nottingham, Kolby and Maniar, Ayush and Robinson, Max and Lillemark, Hansen and Maheshwari, Mehul and Qin, Lianhui and Ammanabrolu, Prithviraj},
	year         = 2025,
	journal      = {arXiv preprint arXiv:2504.17950}
}

@misc{schwartzGreenAI2019,
	title        = {Green {{AI}}},
	author       = {Schwartz, Roy and Dodge, Jesse and Smith, Noah A. and Etzioni, Oren},
	year         = 2019,
	month        = aug,
	publisher    = {arXiv},
	number       = {arXiv:1907.10597},
	doi          = {10.48550/arXiv.1907.10597},
	urldate      = {2024-12-25},
	eprint       = {1907.10597},
	primaryclass = {cs},
	archiveprefix = {arXiv}
}

@article{snell2024scaling,
	title        = {Scaling llm test-time compute optimally can be more effective than scaling model parameters},
	author       = {Snell, Charlie and Lee, Jaehoon and Xu, Kelvin and Kumar, Aviral},
	year         = 2024,
	journal      = {arXiv preprint arXiv:2408.03314}
}

@inproceedings{hung2024discontent,
  title={The Discontent with Intent Estimation In-the-Wild: The Case for Unrealized Intentions},
  author={Hung, Hayley and Li, Litian and Molhoek, Jord and Zhou, Jing},
  booktitle={Extended Abstracts of the CHI Conference on Human Factors in Computing Systems},
  pages={1--9},
  year={2024}
}
